%% file: neurips_2026.tex
\documentclass{article}

\usepackage[USenglish]{babel}
\usepackage{csquotes}

\usepackage[main,nonatbib,final]{neurips_2026}

\makeatletter
\if@preprint \renewcommand{\@noticestring}{} \fi
\makeatother

\usepackage[utf8]{inputenc} %
\usepackage[T1]{fontenc}    %
\usepackage{url}            %
\usepackage{booktabs}       %
\usepackage{amsfonts}       %
\usepackage{nicefrac}       %
\usepackage{microtype}      %
\usepackage{xcolor}         %
\definecolor{mydarkred}{rgb}{0.6,0,0}
\definecolor{mydarkgreen}{rgb}{0,0.6,0}
\usepackage[colorlinks,
linkcolor=mydarkred,
citecolor=mydarkgreen]{hyperref}
\usepackage{graphicx}
\usepackage{subcaption}
\usepackage{amsmath}
\usepackage{amsthm}
\usepackage{mathtools}
\usepackage{todonotes}
\usepackage{mathrsfs}
\usepackage{nicefrac}
\usepackage{thmtools}
\usepackage{wrapfig}
\usepackage[noend]{algpseudocode}
\usepackage{algorithm}

\usepackage[style=authoryear,natbib=true,uniquelist=minyear,maxbibnames=10,maxcitenames=2,abbreviate=false]{biblatex}
\DeclareNameAlias{sortname}{given-family}
\renewbibmacro{in:}{}

\usepackage[capitalize,noabbrev]{cleveref}

\newcommand{\cX}{\mathcal{X}}

\newcommand{\cZ}{\mathcal{Z}}
\newcommand{\cW}{\mathcal{W}}
\newcommand{\cF}{\mathcal{F}}
\newcommand{\cH}{\mathcal{H}}

\newcommand{\sF}{\mathscr{F}}
\newcommand{\fH}{\mathfrak{H}}
\newcommand{\bR}{\mathbb{R}}
\newcommand{\bX}{\mathbb{X}}
\newcommand{\bY}{\mathbb{Y}}
\newcommand{\bZ}{\mathbb{Z}}
\newcommand{\bE}{\mathbb{E}}
\newcommand{\bP}{\mathbb{P}}
\newcommand{\bQ}{\mathbb{Q}}
\DeclareMathOperator*{\argmax}{arg\,max}

\DeclareMathOperator{\Var}{Var}
\DeclareMathOperator{\MMD}{MMD}
\DeclareMathOperator{\E}{\bE}
\newcommand{\hMMD}[1][]{\widehat{\operatorname{MMD}}{}^{#1}}
\newcommand{\tMMD}[1][]{\widetilde{\operatorname{MMD}}{}^{#1}}
\newcommand{\inner}[2]{\left\langle{#1},{#2}\right\rangle}
\algnewcommand{\Input}{\item[\textbf{Input:}]}
\algnewcommand{\Output}{\item[\textbf{Output:}]}
\algnewcommand{\Parameters}{\item[\textbf{Parameters:}]}

\DeclarePairedDelimiter{\abs}{\lvert}{\rvert}
\DeclarePairedDelimiter{\norm}{\lVert}{\rVert}

\newtheorem{theorem}{Theorem}
\newtheorem{proposition}[theorem]{Proposition}
\newtheorem{lemma}[theorem]{Lemma}

\newtheorem{definition}[theorem]{Definition}

\title{Bet Imaginatively, not Historically in Independent-Data Sequential Testing}

\author{%
  Nathaniel Xu\\
  University of British Columbia\\
  \texttt{xunathan@cs.ubc.ca}
  \And
  Feng Liu\\
  University of Melbourne\\
  \texttt{feng.liu1@unimelb.edu.au}
  \And
  Danica J.\ Sutherland\\
  UBC \& Amii\\
  \texttt{dsuth@cs.ubc.ca}
}

\begin{document}

\maketitle

\begin{abstract}
    Sequential testing avoids some of the many controversies and drawbacks related to traditional batch testing with p-values.
    In particular, the \emph{testing by betting} paradigm has developed rapidly in recent years; it frames the level of evidence against the null as the wealth of a player in a betting game.
    Naturally, how much a player should bet is a key consideration, 
    and should reflect their confidence in what they know about the next observed data point.
    Much current work uses standard betting strategies developed for adversarial problems.
    In the stochastic setting, where the observed data is independent and stationary, these approaches may be suboptimal.
    We propose a simple betting strategy,
    \textit{\textbf{f}ollow the \textbf{l}eave-one-\textbf{o}ut \textbf{e}power} (FLOE),
    that exploits the i.i.d. nature of a data stream
    to estimate the e-power of our current test on the underlying distribution.
    We average not over the historical evidence collected so far,
    as generic online learning-based methods do,
    but instead over the equally-likely possible data sequences we might have seen based on shuffling the data.
    With an additional principled regularization technique,
    this simple drop-in method yields substantial improvements in testing performance on a variety of kernel two-sample and independence testing problems.
    Code is available at \url{https://github.com/xunathan96/floe}.
\end{abstract}

\section{Introduction}
\label{sec:intro}
Hypothesis testing is a fundamental area of statistics and is widely used by practitioners across science, medicine, and other areas as a means for verifying preconceived hypotheses. In the machine learning community, tests have been used for a variety of applications including causal inference \citep{peters2015causalinferenceusinginvariant, heinzedeml2018icp, pfister2019icp, liu2024causal}, fairness \citep{romano2020fairness, ferrara2026sizeadaptive}, domain adaptation
\citep{jang2022covariateshift, kubler2022automl2st}, and detecting machine-synthesized content \citep{lopezpaz2017c2st, sutherland2017generative, liu2020deep, chen2025online, li2025watermark, zhou2026adadetectgpt}. 
Historically, work in hypothesis testing has focused on designing \textit{batch} tests, where the practitioner first collects all available data and then uses it to make a decision. These tests are typically based on the p-value of some test statistic of the underlying data distribution,
asking whether the observed statistic value is very unlikely to occur due to chance under the null hypothesis.
This testing paradigm is relatively simple,
and often bears strong theoretical guarantees.

Batch tests, however, are notoriously prone to subtle misuse. 
As an example, suppose a clinical trial is verifying the efficacy of a new drug. This can be framed as a two-sample testing problem between the physical condition of users that took the drug to those that took a placebo. If the conditions are distributionally identical, the drug is ineffective; we call this the \textit{null hypothesis}. The trial first collects data from 100 participants and runs a batch two-sample test on this data: this test is inconclusive, i.e., it fails to reject the null hypothesis. Believing the reason it failed due to insufficient data, the experimenters repeat the test a month later with 200 participants. This time the test does reject the null. Because batch tests are only valid at \textit{predetermined} sample sizes, however, the act of testing on new data -- even if it is all independent! -- invalidates the second test,
as can be clearly seen if the experimenter repeats this process fifty times \citep{ioannidis2005, simmons2011falsepositive}.

The alternative framework of \textit{sequential} tests directly addresses the above concerns. Rather than collecting all available data at once, these tests are designed to be evaluated each time new data is observed. This approach is naturally more data efficient, as the test can be immediately stopped once enough ``evidence'' is collected. This evidence is measured by what are known as \textit{e-values} \citep{vovk2021evalues, shafer2021testing, ramdas2024evalues} -- nonnegative variables whose expectation under the null hypothesis is at most one -- and are in many ways complementary to classical p-values.
Perhaps the biggest advantage is that certain sequences of e-values are naturally closed under multiplication, giving an easy way to aggregate evidence across experiments; this forms the basis of \textit{testing by betting} \citep{shafer2021testing}.

The testing by betting paradigm establishes an equivalence between gathering evidence against the null hypothesis and the wealth of a player in a betting game. This game is constructed so that, if the null hypothesis is true, the game must be fair and the player's wealth is not expected to grow. How much wealth the player has is then directly a measure of how much evidence against the null hypothesis has been accumulated. Thus, a key aspect of this approach is designing betting strategies that maximize the growth rate of a player's wealth. 
Many of the betting strategies used in practice \citep{orabona2016coin, podkopaev2023seq, podkopaev2023seqhsic, chugg2023auditing, chen2025online}, and in particular the kernel sequential tests of \citet{shekar2023seqmmd}, are developed for adversarial online problems where the data is potentially chosen by an adversary. 
However, in the stochastic regime where data is i.i.d., these strategies fail to take advantage of this additional stationarity
-- and many tests in this area \emph{do} assume the data is i.i.d. For the betting strategies that do \citep{waudbysmith2022estimatingmeansboundedrandom, wang2022ebacktracking, pandeva24_deep}, most are based on estimating the expected growth rate, or \textit{e-power}, of the player's wealth.

\textbf{Contributions.}
Early on in the betting game, our hypothesis remains uncertain and so e-power is low.
As we observe more data, we narrow in on a prediction strategy that lets us more accurately assess our hypothesis (e.g. a stronger test statistic that differentiates between observed distributions); this improved prediction hopefully increases the e-power. %
For accurate betting, we should therefore make sure to estimate the e-power based on the \emph{current} statistic, 
not as an average over \textit{previous} statistics.

We do this by exploiting the exchangeability of the data stream
to simulate many equally-likely data sequences; each of these sequences form leave-one-out estimates of the current test statistic, which we then use to estimate the e-power of the subsequent step. This approach leads to better e-power approximations, and thus serves as a much better guide for betting. 
We also prove that this betting strategy eventually recovers the asymptotically optimal bet amount.

Unfortunately, since early e-power estimates (based on scarce data) tend to be noisy, this can lead to overly aggressive early bets.
We thus also propose a regularization strategy
based on the uncertainty in our e-power estimates,
which adaptively allows more aggressive betting in easier problems
but is more conservative in highly uncertain problems.
Adding this regularization
makes our tests more powerful than prior betting strategies
on a variety of real and synthetic testing problems.

\section{Background}
\label{sec:background}
\textbf{Two-Sample Testing.} The batch two-sample testing problem involves determining whether two sets of observed samples $\bX_m:=(X_1,...,X_m) \in \cX^m$ and $\bY_n:=(Y_1,...,Y_n) \in \cX^n$ are drawn from the same distribution. That is, we test the null hypothesis $\fH_0:\bP=\bQ$ against the alternative hypothesis $\fH_1:\bP\ne\bQ$. A hypothesis test $\Delta$ is a $\{0,1\}$-valued function of the data that rejects the null hypothesis when $\Delta(\bX_m,\bY_n)=1$, and does not reject otherwise. Since the samples are random we can never conclude either hypothesis with certainty, and so statistical tests are designed to control the rate of false positives (type-I error) while minimizing the rate of false negatives (type-II error). We say a test is controlled with level $\alpha \in (0,1)$ if its type-I error rate is at most $\alpha$, i.e. $\Pr_{\fH_0}( \Delta(\bX_m,\bY_n) = 1 ) \le \alpha$. Similarly, a test has type-II error rate of $\beta\in(0,1)$ if $\Pr_{\fH_1}(\Delta(\bX_m,\bY_n)=0) \le \beta$. Among all level-$\alpha$ tests, the one with lowest type-II error $\beta$ is considered the most powerful; $1-\beta$ is called the test power. The probability subscripts used above indicate whether samples are drawn under the null or alternative hypotheses, and similar notation will be used for expectations; when none is written, we assume the distribution is under the alternative regime. 

\textbf{Maximum Mean Discrepancy.} The \emph{Maximum Mean Discrepancy} (MMD) \citep{gretton12a_mmd} is a measure of distance between distributions, and can be viewed as an integral probability metric over functions in a reproducing kernel Hilbert space (RKHS) $\cF$ \citep{Aronszajn1950rkhs}:
\begin{align}
    \label{eq:mmd_ipm}
    \MMD^2(\bP,\bQ) := \sup_{f\in\cF, \|f\|\le 1} \bE_{X,Y\sim\bP\times\bQ}[ f(X) - f(Y) ].
\end{align}
$\cF$ is uniquely characterized by a symmetric positive definite kernel $k:\cX\times\cX\to\bR$, and if this kernel is characteristic \citep{sriperumbudur10a} then $\MMD^2(\bP,\bQ) = 0$ iff $\bP = \bQ$. Gaussian kernels $k(x,y) = \exp(-\frac{1}{2}\|x-y\|^2 / \gamma^2)$ with lengthscale $\gamma>0$ are a common such example.

When the MMD is computed between two empirical distributions $\hat\bP_m$ and $\hat\bQ_n$, we get a biased but consistent estimate of the population value. If the sample sizes are equal, i.e. $m=n$, this estimate is a one-sample V-statistic on the paired samples $\bZ_n=(Z_1,...,Z_n)$ with $Z_i = (X_i,Y_i)$ and given by
\begin{align}
    \label{eq:mmd_vstat}
    \hMMD[2]_b(\bZ_n) = \frac{1}{n^2} \sum_{i=1}^n \sum_{j=1}^n h(Z_i,Z_j),
\end{align}
where we define $h(Z,Z') := k(X,X') - k(X,Y') - k(Y,X') + k(Y,Y')$ to be the symmetric V-statistic kernel function. This quadratic-time estimator is straightforward to compute and is a popular statistic for non-parametric tests. In particular, a batch two-sample test can be constructed with finite-sample validity by comparing the $p$-values of the MMD statistic under permutations of the data \citep{hemerik2014perm}. For any fixed sample size $n$, the test is controlled with level at most $\alpha$.
Like typical $p$-value tests, however,
they are \emph{not} robust to phenomena such as $p$-hacking;
$\Pr_{\fH_0}( \max_{n>0} \Delta(\bZ_n) = 1)$ is likely to greatly exceed $\alpha$.

\textbf{Sequential Tests.}
A desirable alternative to batch tests are \emph{anytime valid tests}, which have type-I error controlled uniformly across sample sizes. These \textit{sequential tests} and are particularly suited for online testing problems where data is continuously collected. We define a sequential test based on its stopping time $\tau := \inf\{ t \ge 1: \Delta(\bZ_t) = 1 \}$, the smallest sample size $t$ such that a batch test on samples $Z_1,...,Z_t$ rejects the null hypothesis.
Here we require time-uniform control:
$\Pr_{\fH_0}(\tau < \infty) \le \alpha$ for some prescribed level $\alpha\in(0,1)$. The sequential test $\tau$ is said to be \emph{consistent} if $\Pr_{\fH_1}( \tau < \infty ) = 1$, that is, the process rejects almost surely under the alternative hypothesis.

Recently, the \textit{testing by betting} principle \citep{shafer2021testing} has been popularized as a blueprint for designing anytime valid sequential tests \citep{shekar2023seqmmd,podkopaev2023seq,pandeva24_deep}. The idea is to define a betting game where at every step $t$ we bet on predictions whose odds are sub-fair under the null hypothesis. Then the cumulative wealth at any time characterizes the the amount of evidence against the null up to that point. We formalize this game as follows: a player starts with initial wealth $\cW_0=1$. At step $t$, they select a prediction function $\nu_t:\cZ\to [-1,\infty)$ such that $\bE_{\fH_0}[\nu_t(Z)\mid \sF_{t-1}] \le 0$, where $\sF_{t-1}=\sigma(\{Z_1,...,Z_{t-1} \})$ is the natural filtration, i.e.\ ``everything seen before time $t$.'' They also select an $\sF_{t-1}$-measurable bet $\lambda_t\in[0,1)$. Then, after observing sample $Z_t$, their wealth is updated according to
\begin{align}
    \label{eq:wealth_process}
    \cW_0 = 1,
    \qquad
    \cW_t = \cW_{t-1}\left( 1 + \lambda_t \nu_t(Z_t)  \right).
\end{align}
This wealth process forms a \textit{test supermartingale} under the null \citep{ramdas2024evalues}.
A supermartingale is a random process where for all $t$, $\E[ \cW_t \mid \sF_{t-1} ] \le \cW_{t-1}$
and $\E \abs{\cW_t} < \infty$;
a test supermartingale adds the conditions that $\cW_0 = 1$ and that $\cW_t$ is almost surely nonnegative.
A level-$\alpha$ sequential test for $\fH_0$ against $\fH_1$ is then defined by the stopping time 
\begin{align}
    \label{eq:stopping_time}
    \tau = \inf\{ t \ge 1: \cW_t \ge 1/\alpha \},
\end{align}
and anytime validity follows from Ville's inequality, a time-uniform generalization of Markov's inequality for nonnegative supermartingales.

\textbf{Prediction and Betting Strategies.}
The power of this test depends on the strategies for selecting the predictors $\nu_t$ and bets $\lambda_t$. In the kernel two-sample sequential test of \citet{shekar2023seqmmd}, they propose using the prediction function $\nu_t(Z) = f_t(X) - f_t(Y)$ for some $\sF_{t-1}$-measurable RKHS function $f_t$; this ensures $\bE_{\fH_0}[f_t(X) - f_t(Y) \mid \sF_{t-1}] = 0$. To get positive drift under the alternative, a natural choice for $f_t$ is the empirical risk minimizer (ERM), the RKHS function that maximizes \cref{eq:mmd_ipm} for empirical distributions $\hat\bP_{t-1}$ and $\hat\bQ_{t-1}$:
\[
    f(t) \propto \E_{X \sim \hat\bP_{t-1}}[ k(X, t) ] - \E_{Y \sim \hat\bQ_{t-1}}[ k(Y, t) ]
.\]
If the RKHS kernel $k$ satisfies $k(x,x) \le K$, then $f_t$ is normalized to be in a ball of radius $1/(2\sqrt{K})$ so that $f_t(X) \in [-1/2,1/2]$; this ensures the payoffs $\nu_t(Z)$ are in $(-1,1)$.
The bets can also be chosen in a similar manner, where we set $\lambda_t$ to be the scalar in $(-1,1)$ that maximizes the log-wealth at time $t-1$; this is known as GRAPA \citep{waudbysmith2022estimatingmeansboundedrandom} in the literature.

In the context of online learning, both of the aforementioned approaches are \textit{follow-the-leader} (FTL) strategies because they use what would have been the optimal value in the previous round. Instead, \citet{shekar2023seqmmd} opt to use different online strategies, namely \textit{online gradient descent} (OGD) \citep{zinkevich2003OnlineCP} for selecting the critic $f_t$, and \textit{online Newton step} (ONS) \citep{hazan2007ons} for the bet $\lambda_t$. Their motivation for doing so is largely theoretical, as OGD and ONS are shown to respectively achieve $O(\sqrt{T})$ and $O(\log T)$ regret even in the adversarial online setting, whereas FTL strategies may not exhibit sub-linear regret. 

Another popular online strategy with strong theoretical guarantees is the $\mu$-weighted \textit{universal portfolio} algorithm (UP) of \citet{cover1996}, which was originally motivated as an investment strategy for rebalancing $m$ stocks. The authors show that this algorithm achieves minimax-optimal regret bounds when used with a $\mathrm{Dirichlet}(1/2,...,1/2)$ prior. Since selecting scalar bets in $[0,1]$ is equivalent to rebalancing a two-stock portfolio, this simplifies to a $\mathrm{Beta}(1/2,1/2)$ prior, and numerical integration becomes a practical approach to computing the bets.

The OGD, ONS, and UP algorithms all apply to the general online setting where data could be adversarially selected. In the stochastic regime, however, a test based on FTL payoff and betting strategies is still consistent even if its worst-case adversarial regret is poor. As such, it is not clear which strategy performs better in practice.

\section{Proposed Method}
\label{sec:method}
\subsection{Oracle Test}
We start by constructing the oracle sequential two-sample test for witness functions in a bounded reproducing kernel Hilbert space,
similar to \citet{shekar2023seqmmd}.
This test assumes complete knowledge of the underlying data distribution,
and involves separate optimal strategies for learning the payoffs and bets.  
Henceforth, we assume bounded kernels with $\sup_{x} k(x, x) \le K$ and $\inf_{x, y} k(x, y) \ge 0$.

The \textit{witness function} $f^*$ is the member in $\cF$ that achieves the supremum in \cref{eq:mmd_ipm}; $f^*$ ``witnesses'' the difference between $\bP$ and $\bQ$.
For a version of \eqref{eq:mmd_ipm} 
normalized to a radius-$1\!/\!\sqrt{2K}$ ball, 
which just yields the value $\MMD(\bP, \bQ) / \sqrt {2K}$,
this maximizer is simply
\begin{align}
    \label{eq:witness_rkhs}
    f^*
    := \argmax_{f\in\cF, \|f\|\le \frac{1}{\sqrt{2K}} } \bE[f(X) - f(Y)] 
    = \frac{\mu_\bP - \mu_\bQ}{\sqrt{2K} \| \mu_\bP - \mu_\bQ \|},
\end{align}
where $\mu_\bP:=\bE_\bP[k(\cdot,X)]$ and $\mu_\bQ:=\bE_\bQ[k(\cdot,Y)]$ are known as the \textit{mean embeddings} of $\bP$ and $\bQ$.
We define the \textit{witness predictor} as the function $\nu^*(x,y) := f^*(x) - f^*(y)$, and note that its values are necessarily bounded by one since
\[
    \abs{\nu^*(x,y)} 
    = \abs*{ \inner{f^*}{k(\cdot,x) - k(\cdot,y)} }
    \le \norm{f^*} \norm{k(\cdot,x) - k(\cdot,y)}
    \le \frac{1}{\sqrt {2K}} \sqrt{2K}
    = 1
.\]
Its prediction at a paired sample $z=(x,y)$ is, with $h(z,z') = k(x,x') - k(x,y') - k(y,x') + k(y,y')$,
\begin{gather}
    \label{eq:witness_predictor}
    \nu^*(z) = f^*(x) - f^*(y) = \frac{\bE_Z[h(z,Z)]}{\sqrt{2K\,\bE[h(Z,Z')]}}
.\end{gather}
We denote this predictor evaluated at time $t$ on the sample $Z_t$ to be $V_t^* := \nu^*(Z_t)$, and call this the \textit{witness payoff} at $t$. One can confirm that $\bE_{\fH_0}[V^*_t \mid \sF_{t-1}] = 0$, either by noting that $f^*$ does not depend on the data, or by identifying $\bE_{\fH_0}[h(Z_t,Z')] = \MMD^2(\bP,\bP) = 0$.

As for the bets, a natural approach is to, at every step $t$, pick the value in $[0,1)$ that maximizes the expected growth in wealth.
(Since $V_t^* \ge -1$, the restriction $\lambda < 1$ ensures that $1 + \lambda^* V_t^* > 0$ almost surely.)
This strategy is essentially the \textit{Kelly criterion} for portfolio optimization:
\begin{align}
    \label{eq:kelly_criterion}
    \lambda_t^* \in \argmax_{\lambda\in[0,1)} \bE[\log(1 + \lambda V_t^*) \mid \sF_{t-1}].
\end{align}
The $\bE[\log(1+\lambda V_t^*) \mid \sF_{t-1} ]$ term is the expected log-growth of the wealth at time $t$ and is also known as the \textit{e-power}.
In the i.i.d. setting, the witness payoffs $V_t^*$ are independent of the past, and so this reduces to an unconditional expectation; thus $\lambda_t^*$ is the same for all time $t$ and is simply denoted as $\lambda^*$. Putting everything together, we construct the oracle wealth process as 
\begin{align}
    \label{eq:wealth_process_oracle}
    \cW_0^* = 1,
    \qquad
    \cW_t^* = \cW_{t-1}^*( 1 + \lambda^* V_t^* ),
\end{align}
and the oracle sequential test as the stopping time $\tau := \inf\{ t \ge 1: \cW_t^* \ge 1/\alpha \}$. This test exhibits time-uniform control with level $\alpha$, and is known to be consistent under the alternative hypothesis. 

\subsection{Practical Test}
\label{sec:pratical_test}

Although the oracle test is informative, it is impossible to construct since we do not know the underlying data distribution. Instead, we define our prediction and betting strategies based on estimates of the witness function.
To ensure such strategies are valid, we require them to be predictable, i.e.\ $\sF_{t-1}$-measurable; this way the wealth process of \eqref{eq:wealth_process} forms a test supermartingale.

One standard approach is to compute the empirical risk minimizer for \eqref{eq:mmd_ipm} between the empirical distributions $\hat\bP_{t-1}$ and $\hat\bQ_{t-1}$.
This gives us a predictable critic $f_t$,
and consequently the predictor $\nu_t(z) := f_t(x) - f_t(y)$ and payoff $V_t := \nu_t(Z_t)$.
We will stick with this approach.
In the stochastic setting,
there is little reason to choose otherwise;
by symmetry,
it cannot help to use another fixed weighting of each data point.
Although Stein-style shrinkage estimators of the kernel mean exist \citep{krikamol:shrinkage},
the ERM estimator gives minimax-optimal estimates of the MMD \citep{ilya:mmd-minimax}.
It is worth noting as a pointer for future work,
however,
that this choice does not necessarily maximize the e-power of the test.

If we decide to use the standard wealth process $\cW_t = \cW_{t-1}( 1 + \lambda_t \nu_t(Z_t) )$,
it remains to choose a predictable bet $\lambda_t$.
Ideally, we would choose $\lambda_t$ according to the Kelly criterion \eqref{eq:kelly_criterion}
for payoffs based on the current critic at time $t$:
\begin{equation} \label{eq:kelly-nu_t}
    \lambda_t \in \argmax_{\lambda \in [0, 1)} \E\bigl[ \log(1 + \lambda \nu_t(Z) ) \mid \sF_{t-1} \bigr]
.\end{equation}
However, since our bet must be predictable
and $\nu_t$ is based on \emph{all} of the available data (i.e., $Z_1,...,Z_{t-1}$)
we have no held-out samples with which to estimate this e-power.
The GRAPA betting strategy \citep{waudbysmith2022estimatingmeansboundedrandom},
which stands for \emph{growth rate adaptive to the particular alternative},
therefore estimates the e-power
based on maximizing
$\frac{1}{t-1} \sum_{\ell=1}^{t-1} \log (1 + \lambda \nu_\ell(Z_\ell))$,
which is equivalent to maximizing the wealth at time $t-1$
for a process using the fixed bet $\lambda$
(a ``follow-the-leader'' strategy).
In the limit as $\ell \to \infty$,
we will typically have $\nu_\ell \to \nu^*$
and $\lambda$ will converge to $\lambda^*$ of \eqref{eq:kelly_criterion}.

As previewed in \cref{sec:intro}, however,
the key insight of our work is that $\frac{1}{t-1}\sum_{\ell<t} \log(1 + \lambda \nu_\ell(Z_\ell))$ is \emph{not} necessarily a good estimate of the distribution of returns for $V_t = \nu_t(Z_t)$.
Hopefully our predictor $\nu_t$ has gotten better as we observe more data,
but GRAPA weights the historical return of $\nu_2$ equally to how it weights $\nu_{t-2}$!
We suggest a simple alternative:
while we do not have access to fresh data on which to estimate $\E \log(1 + \lambda \nu_t(Z_t))$,
in the stochastic setting we can use a \emph{leave-one-out estimate} of the e-power.
That is, rather than plugging in an average over the history of our actual critic functions,
we can plug in an average over $t-1$ hypothetical sequences
in which the last point we observed was $Z_1$, $Z_2$, \dots, or $Z_{t-1}$.
Because our data is exchangeable, this is an unbiased estimator for the e-power of $\nu_{t-1}$,
and hence a good estimate for the e-power of $\nu_t$,
the actually relevant quantity.
We now detail how this is done.

\textbf{Critic construction.}
Write $[t] = \{1, 2, \dots, t\}$, so that $[t] \setminus \{\ell\}$ is the set of all indices up to $t$ but excluding $\ell$.
Let $\hat{\bP}_{[t] \setminus \{\ell\}}$ and $\hat{\bQ}_{[t] \setminus \{\ell\}}$ be the empirical distributions based on $Z_i$ for $i \in [t] \setminus \{\ell\}$. Define the \textit{leave-one-out (loo) critic} $f_{[t] \setminus \{\ell\}}$ as the norm-$1\!/\!\sqrt{2K}$ maximizer of the MMD between empirical distributions $\hat{\bP}_{[t] \setminus \{\ell\}}$ and $\hat{\bQ}_{[t] \setminus \{\ell\}}$, 
and the corresponding predictor
$\nu_{[t] \setminus \{\ell\}}(x,y) = f_{[t] \setminus \{\ell\}}(x) - f_{[t] \setminus \{\ell\}}(y)$. It is straightforward to show that this predictor evaluated on a paired sample $z=(x,y)$ is
\begin{align}
    \label{eq:loo_predictor}
    \nu_{[t]\setminus \{\ell\}}(z)
    = \frac{\sum_{i \in [t] \setminus \{\ell\}} h(z, Z_i) }{ \sqrt{ 2K \sum_{i \in [t] \setminus \{\ell\}} \sum_{j \in [t] \setminus \{\ell\}} h(Z_i,Z_j) } }.
\end{align}
When we evaluate a loo-predictor on its left-out sample,
we obtain the \textit{leave-one-out (loo) payoff} $V_\ell^{(t)} := \nu_{[t] \setminus \{\ell\}}(Z_\ell)$.
It is worth mentioning that
since of course $[t] \setminus \{t\} = [t-1]$,
$\nu_{[t] \setminus \{t\}}$ is exactly the standard ERM predictor $\nu_t$,
and $V_t^{(t)}$ the ERM payoff $V_t$.
Of the leave-one-out predictors,
only $\nu_t = \nu_{[t] \setminus \{t\}}$ is $\sF_{t-1}$-measurable,
and hence only $V_t$ is a valid payoff at time $t$.

\textbf{Betting strategy.}
Recall that the oracle betting strategy aims to maximize the e-power of the wealth process at time $t$. Although the exact e-power is impossible to determine, we can construct a $\sF_{t-1}$-measurable estimate of it from the loo-payoffs at time $t-1$. Our betting strategy is the maximizer of this estimate in the interval $[0,1-\epsilon]$ for some small $\epsilon>0$, i.e.
\begin{align}
    \label{eq:loo_bet}
    \lambda_t := \argmax_{\lambda\in[0,1-\epsilon]} \frac{1}{t-1} \sum_{\ell=1}^{t-1}\log(1 + \lambda V_{\ell}^{(t-1)}).
\end{align}
We call this betting strategy \textit{follow-the-loo-e-power} (FLOE).
As the e-power estimate is strictly\footnote{Provided at least one payoff is non-zero, which is almost surely true for non-degenerate distributions.} concave with respect to $\lambda\in[0,1)$, a unique maximizer exists.
This one-dimensional optimization problem can be done via bisection to $\varepsilon$ accuracy in $\log_2(\varepsilon)$ steps;
five or ten steps are more than enough for practical purposes.
Moreover, computing the e-power estimate at time $t$ can be done in $O(t)$ cost as long as we saved the kernel matrix $H_{ij}:=h(Z_i,Z_j)$ of the previous step.
This betting strategy also converges to the Kelly criterion bet (\cref{prop:loo-epower-conv}).

Putting everything together,
we use bets $\lambda_t$ chosen according to \eqref{eq:loo_bet}
and ERM payoffs $V_t = \nu_t(Z_t)$,
where $\nu_t$ is the norm-$1/\sqrt {2K}$ witness predictor between $\hat\bP_{[t-1]}$ and $\hat\bQ_{[t-1]}$.
The test martingale is
\begin{align}
    \label{eq:wealth_process_practical}
    \cW_0=1,
    \qquad
    \cW_t = \cW_{t-1}( 1 + \lambda_t V_t )
.\end{align}
The corresponding sequential level-$\alpha$ test is the stopping time $\tau:=\inf\{ t \ge 1: \cW_t \ge 1/\alpha \}$, which under the null is finite with probability no greater than $\alpha$.

\textbf{Test Consistency.}
Although the test is guaranteed to be anytime valid with level-$\alpha$, we would also like it to be consistent: it should eventually reject the null under the alternative. If not, one could simply construct a valid, but uninformative, sequential test like $\cW_t = 1$. The following theoretical result confirms the wealth process of \cref{eq:wealth_process_practical} is consistent while maintaining finite-sample type-I control. Its proof is provided in \cref{app:test_consistency}.

\begin{restatable}{theorem}{maintheorem}
    \label{thm:consistency}
    If $\tau := \min\{t \ge 1: \cW_t \ge 1/\alpha\}$ with $\cW_t$ from \eqref{eq:wealth_process_practical} with a characteristic kernel,
    then
    \[
        \Pr_{\fH_0}(\tau < \infty) \le \alpha 
        \quad
        \text{and}
        \quad
        \Pr_{\fH_1}(\tau < \infty) = 1
    .\]
\end{restatable}

\subsection{Bet Regularization}

\begin{wrapfigure}{r}{0.4\linewidth}
    \centering
    \includegraphics[width=\linewidth]{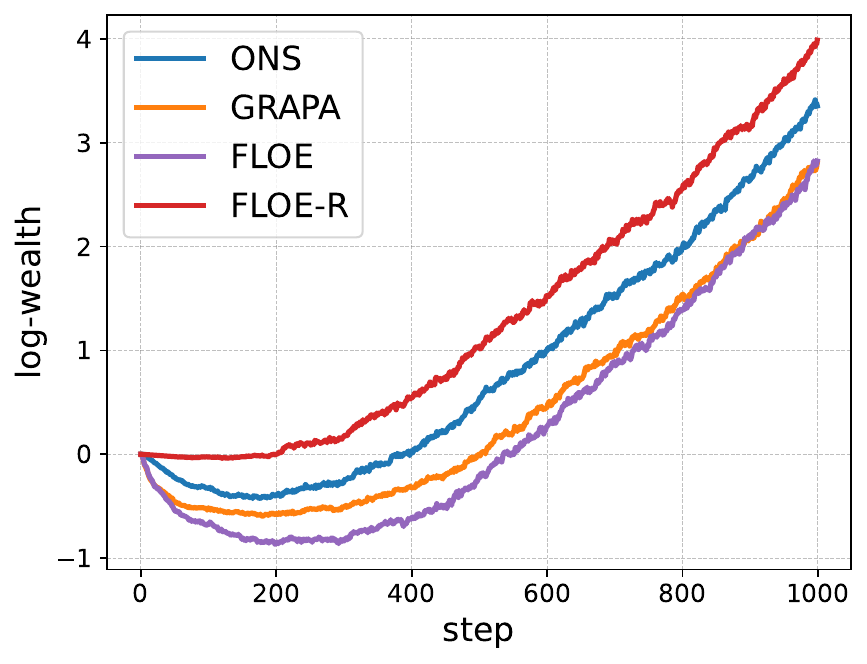}
    \caption{Wealth of several betting strategies (ONS, GRAPA, FLOE) for the x-Normal problem described in \cref{sec:experiments}. Poor initial estimates often lead to negative returns, whereas a regularized betting strategy (FLOE-R) minimizes losses by betting more conservatively at the start.}
    \label{fig:bet_reg_motivation}
\end{wrapfigure}

The finite sample performance of any betting strategy is subtly tied to the uncertainty of predictor outputs. At sufficiently large $t$, we can expect the ERM predictors to be good approximations of the witness predictor, and so we can be confident their evaluated payoffs are accurate measures of evidence against the null hypothesis. On the other hand, we should be more skeptical of payoffs when $t$ is small, as predictions are less accurate in this regime. An ideal betting strategy should be aggressive when we are confident in the payoffs and conservative when we are not. 
\cref{fig:bet_reg_motivation} illustrates what happens when uncertainty is not encoded into the betting strategy. When $t$ is small, the ERM payoffs are more frequently negative, which causes wealth to initially decrease. This phenomenon motivates using an uncertainty-based regularizer for the bets.

There are several ways we can quantify this uncertainty. Perhaps the most direct approach is with high probability bounds on either the predictor $\nu_t$ or its payoff $V_t$. But, while informative, these bounds alone do not justify particular betting values. %
We address this issue by defining a regularization paradigm based on the expected return  $\bE[\lambda V_t \mid \sF_{t-1}]$, which directly connects to the choice of bets.

\begin{definition}
    \label{def:bet_reg}
    Let $\alpha,\beta \in (0,1)$,
    and $\bZ_{t-1} = (Z_1, \dots, Z_{t-1})$.
    We call $\Lambda_{t}^{(\alpha,\beta)}(\bZ_{t-1}) \subseteq [0,1)$ an $(\alpha,\beta)$-regularized set of bets for the ERM predictor at time $t$ if
    \begin{align*}
        \Pr_{\bZ_{t-1}} \left( \inf_{\lambda\in \Lambda_t^{(\alpha,\beta)}(\bZ_{t-1})} \bE[ \lambda V_{t} \mid \sF_{t-1} ] > - \alpha \right) \ge 1 - \beta
    .\end{align*}
\end{definition}

An $(\alpha,\beta)$-regularized set of bets is an interval of bets that guarantees an expected return of at least $-\alpha$ with probability no less than $1-\beta$.
This definition inherently accounts for the uncertainty of the prediction function, as we need to be confident the expected profit will not drop too low. It is worth pointing out that bets in $\Lambda_t^{(\alpha,\beta)}$ also indirectly regularize the e-power $\bE[\log(1 + \lambda V_t) \mid \sF_{t-1}]$ through \cref{lem:indirect-epower-reg}, though this bound is certainly not sharp.

The next theorem (proved in \cref{app:bet_reg}) constructs one possible $(\alpha, \beta)$-regularized set. It is based on a martingale version of a Bennett-type concentration inequality, which uses variance information to provide sharper bounds. In this case, it uses the variance of the first-order projection of the V-statistic kernel, which we denote by $\zeta(\bP\times\bQ) := \Var_Z[ \bE_{Z'}[ h(Z,Z') ] ]$.
This term is essentially the asymptotic variance of the MMD estimator in \cref{eq:mmd_vstat}; estimating this is required for several notable kernel two-sample testing approaches \citep{sutherland2017generative, liu2020deep,sutherland:unbiased-mmd-variance, wei2026maximum}. The typical biased estimate is
\begin{align}
    \label{eq:mmd_var}
    \hat{\zeta}_b(\bZ_t) := \frac{1}{t}\sum_{i=1}^{t} \left( \frac{1}{t} \sum_{j=1}^t h(Z_i,Z_j) - \frac{1}{t^2} \sum_{k=1}^t \sum_{j=1}^t h(Z_k,Z_j) \right)^2.
\end{align}

\begin{restatable}{theorem}{theorembet}
    \label{thm:bet_reg}
    Let $\bZ_t = (Z_1,...,Z_t)$ be i.i.d. samples from $\bP\times\bQ$ and suppose we have tail bounds on the estimates in \cref{eq:mmd_vstat} and \cref{eq:mmd_var},
    \begin{align*}
        \Pr\left( \hMMD[2]_b(\bZ_t) - \MMD^2(\bP,\bQ) \le  \varepsilon_t^{(1)} \right) \ge 1 - \delta_1 \\
        \Pr\left( \zeta(\bP \!\times\! \bQ) - \hat\zeta_b(\bZ_t) \le  \varepsilon_t^{(2)} \right) \ge 1 - \delta_2
    ,\end{align*}
    with non-negative errors $\varepsilon_t^{(1)} = o(1)$ and $\varepsilon_t^{(2)} = o(1)$. For brevity, we omit arguments $\bP$, $\bQ$, $\bZ_t$ and define the penalized estimates $\tMMD[2]_b := \hMMD[2]_b - \varepsilon_t^{(1)}$ and $\widetilde{\zeta}_b := \hat{\zeta}_b + \varepsilon_t^{(2)}$. Then for all $\alpha\in(0,1)$
    \begin{align*}
        \lambda_{t+1} \in \left[ 0, \min\left\{ 1, \frac{ \alpha\, \sqrt{2K} \, \hMMD_b }{ \left[ \frac14 {\widetilde{\zeta}_b} \phi_+^{-1} \left( \frac{16 \log 1/\delta_0}{ t \,\widetilde{\zeta}_b } \right) - \tMMD[2]_b \right]_+ } \right\} \right)
    \end{align*}
    is a $(\alpha, \delta_0 + \delta_1 + \delta_2)$-regularized set of bets for the ERM predictor at time $t+1$, where $\phi_+^{-1}$ is the positive inverse of $\phi(u) = (1+u)\log(1+u) - u$.
\end{restatable}

The first term in the denominator is expected to decrease with time, causing the bet interval to increase. We illustrate an example of this interval for a real wealth process in \cref{fig:bet_interval} of the appendix,
and point the reader back to \cref{fig:bet_reg_motivation} for its impact on the wealth. 
This regularization can also be applied to any betting strategy, not just for FLOE, as long as we use an ERM prediction strategy. We find it helps FLOE more than ONS, however, as the latter naturally takes smaller increments. Comparisons between the two are given in \cref{app:ons_reg}.

Finally, we emphasize that this regularization does not increase asymptotic computational complexity. Computing its bound at time $t$ can be done in $O(t)$ time using the saved kernel gram matrix $H_{ij}$ from the previous step. Cumulatively over $T$ steps, this amounts to $O(T^2)$ complexity in both compute and memory, the same as any kernel sequential test based on ERM or OGA prediction strategies. The complete kernel sequential test with this regularization procedure is outlined in \cref{alg:seq_test}.

\begin{algorithm}[hbt!]
    \caption{Sequential test with the regularized FLOE betting strategy}
    \label{alg:seq_test}
    \begin{algorithmic}
        \Input Data stream $Z_1,Z_2,...,Z_T$ sampled independently from $\bP\times\bQ$. 
        \Parameters Test level $\eta\in(0,1)$. Regularization strength $\alpha, \delta_0, \delta_1, \delta_2  \in (0,1)$.
        \Output $\{0,1\}$ where $0$ indicates `do not reject' and $1$ is `reject' the null hypothesis $\fH_0:\bP = \bQ$.
        \State $\cW_0 \gets 1$
        \For{$t=1,2,...,T$} 
            \State Compute ERM payoff $V_t = \nu_t(Z_t)$
            \State Compute bet $\lambda_t$ according to \cref{eq:loo_bet}
            \State Construct the interval $\Lambda_t^{(\alpha, \delta_0 + \delta_1 + \delta_2)}$ described in \cref{thm:bet_reg}
            \State $\lambda_t \gets  $ Project $\lambda_t$ onto $\Lambda_t^{(\alpha, \delta_0 + \delta_1 + \delta_2)}$
            \State $\cW_t \gets \cW_{t-1} ( 1 + \lambda_t V_t ) $
            \If{ $\cW_t \ge 1/\eta$ }
                \State\Return 1
            \EndIf
        \EndFor
        \State\Return 0
    \end{algorithmic}
\end{algorithm}

\textbf{Practical Considerations.}
\Cref{thm:bet_reg} gives a general way to regularize the bets, but it also depends on how tight the tail bounds on $\hMMD[2]_b(\bZ_t)$ and $\hat{\zeta}_b(\bZ_t)$ are. If these bounds are too loose, then the regularization set becomes too conservative, i.e. the interval remains small even after large time $t$. 

For the MMD estimator this is not a major issue, as there exist $O(t^{-1/2})$ error bounds with reasonable constants; for example, we know $\varepsilon_t^{(1)} = 2/t + 4 \sqrt{[\log 1 / \delta_1]/t}$ from a straightforward application of McDiarmid's inequality. The variance estimator $\hat{\zeta}_b$ is more problematic. 
Not only do we need a practical bound with small enough constants, but this bound should be on the square root of the estimator $\surd\hat{\zeta}_b$, because $\hat{\zeta}_b \phi_+^{-1}(1/\hat{\zeta}_b)$ simplifies to a leading term depending on $\surd\hat{\zeta}_b$ once we take the Taylor expansion. A peeling argument or entropy method could be used to do this, but neither would give practical bounds. Therefore, we omit penalizing $\hat{\zeta}_b$ in practice and simply use the unpenalized variance estimate when constructing the regularized set. Although this sacrifices any finite-sample guarantees, we expect it to still approximate the theoretical interval well for large enough $t$.

Another simplification we employ in practice is to replace $\phi_+^{-1}(u)$ with $h^{-1}(u) := u/3 + \sqrt{2u}$, which is the relaxation used when deriving Bernstein's inequality from Bennett's. This still ensures the correct confidence level of the regularization set, since $\phi_+^{-1}(u) \le h^{-1}(u)$ for all $u \ge 0$.

\section{Experiments}
\label{sec:experiments}
\begin{figure}[hbt!]
  \centering
  \begin{subfigure}{\linewidth}
    \centering
    \includegraphics[width=0.7\linewidth]{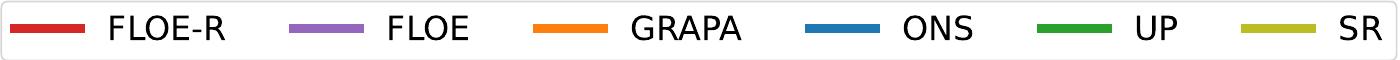}
  \end{subfigure}
  \begin{subfigure}[b]{\linewidth}
    \centering
    \includegraphics[width=\linewidth,trim={0 2cm 0 0},clip]{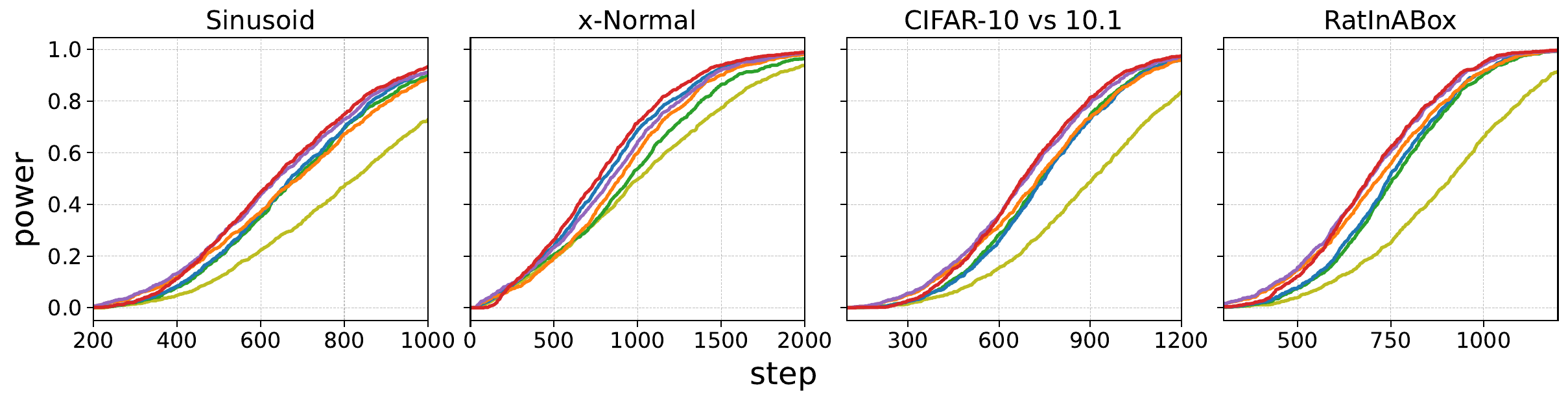}
  \end{subfigure}
  \begin{subfigure}[b]{\linewidth}
    \centering
    \includegraphics[width=\linewidth,trim={0 0 0 9mm},clip]{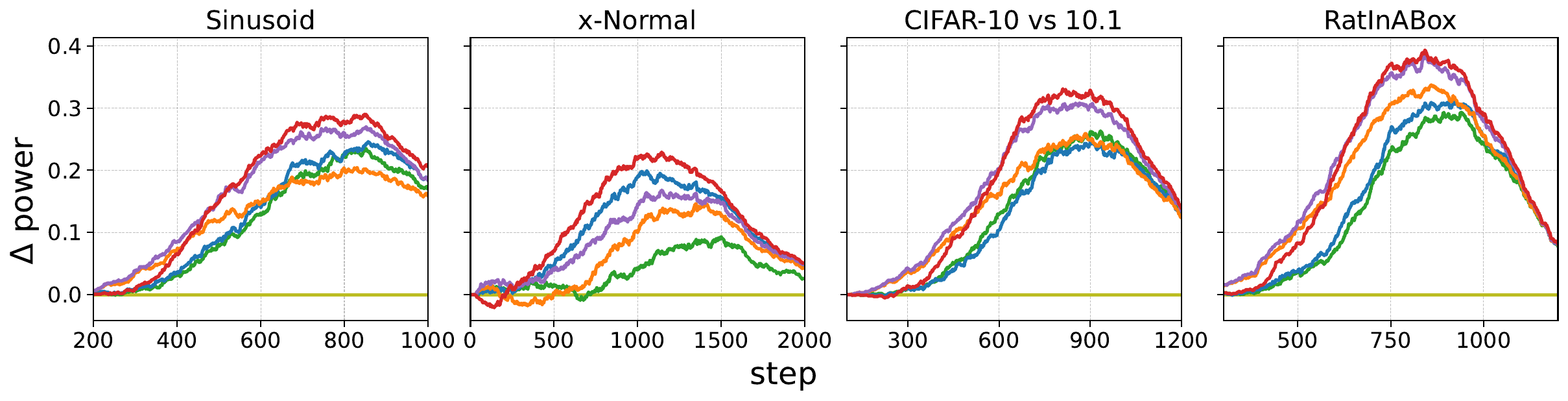}
  \end{subfigure}
  \caption{Mean test power on a variety of datasets; each test is run 1,000 times. The second row subtracts the performance of SR from each point for easier visualization.}
  \label{fig:baselines_floe}
  \vspace{-2em}
\end{figure}

We evaluate our proposed sequential test across a number of synthetic and real-life datasets. Its performance with regularization (FLOE-R) and without (FLOE) is compared to a number of baselines including ONS \citep{hazan2007ons}, GRAPA \citep{waudbysmith2022estimatingmeansboundedrandom}, UP \citep{cover1996, orabona2022tightconcentrationsconfidencesequences} with a Beta($1/2$,$1/2$) prior, and the kernel sequential test of \citet{shekar2023seqmmd} (labelled SR) which uses OGD for the prediction strategy and ONS for the bets. All tests are controlled with level-$0.05$, and the regularization strength for FLOE-R is set to $\alpha=\delta_0=\delta_1=0.1$ while omitting the variance penalty. We use Gaussian kernels for all problems, and select the lengthscale to optimize the performance of ONS; tuning for FLOE(-R) might lead to slightly larger improvements.
Results are illustrated in \cref{fig:baselines_floe}. Further experiment details and additional results on the Blobs and Higgs datasets are given in \cref{app:more-experiments}. 

\textbf{Sinusoid.}
This problem, adapted from \citet{sejdinovic2012hypothesistestingusingpairwise}, involves differentiating between sinusoidal and uniform distributions in a compact space. We sample $\bP$ from the standard uniform on $[-\pi,\pi]^2$, and $\bQ$ according to $\propto 1 + \sin(\omega x) \sin(\omega y)$ on the same support. Higher frequencies $\omega$ produce subtler departures from the uniform distribution, resulting in a harder two-sample problem.
We use $\omega=3.5$ and visualize its density in \cref{fig:density_sinusoid}.

\textbf{x-Normal.}
We distinguish a 2D Gaussian mixture forming an ``X'' shape (\cref{fig:density_hdgm}) from an isotropic Gaussian. Each mixture component of the X is constructed with a positive and negative correlation coefficient of equal magnitude; the lower the correlation, the harder it is to distinguish from a standard Gaussian. We use a correlation magnitude of $0.65$. This problem is essentially the HDGM problem of \citet{liu2020deep} in two dimensions.

\textbf{CIFAR 10 vs 10.1.}
This problem involves discerning between images from the CIFAR-10 \citep{krizhevsky2009learning} and CIFAR-10.1 \citep{torralba2008tinyimages, recht2018cifar10classifiersgeneralizecifar10} test sets. The latter dataset was created entirely from new samples as an alternative test set for models trained on CIFAR-10, and models trained on it generally exhibit a drop in test accuracy. We compare the low-level features of each image obtained from a ResNet-50 model \citep{he2015deepresiduallearningimage} pretrained on ImageNet.

\textbf{RatInABox} \citep{ratinabox}\textbf{.}
We simulate the neural activity of a rat moving inside slightly different boxed environments: one box has a protruding wall, while the other does not. This discrepancy in geometry should result in slightly different place cell (neural cells that are activated in the vicinity of certain locations) firing rates for each environment. We sample from 10 fixed place cells in the rat.

All of the considered tests are significantly better than the test of \citet{shekar2023seqmmd}:
this difference is due to using an ERM prediction strategy
rather than OGD,
since we find the ONS baseline substantially outperforms SR.
As previously mentioned, their motivation for using OGD is theoretical, as there exist known regret bounds for OGD but not for ERM-based strategies.

Among the ERM-based tests, however,
FLOE-R performs as well as, and in most cases better than, every competitor.
Particularly, it improves finite sample performance against ONS and GRAPA by over 5\% for some problems. Even without the regularizer FLOE is enough to beat these other strategies in all benchmarks except x-Normal, and is only slightly worse than FLOE-R in most cases.

In general, including the regularizer sacrifices early power for faster wealth growth, hence higher later power; the largest boost to power we observed is almost 10\% in x-Normal. However, there are situations where including this regularization might be detrimental: if the problem is trivial enough to be solved in only a few steps. In this case, its performance hinges on how sharp the bound in \cref{def:bet_reg} is.

FLOE is also noticeably better than Universal Portfolio, which was shown by \citet{cover1996} to achieve minimax optimal regret against the best in-hindsight constant betting strategy. The optimal strategy, however, is not constant, and so non-constant algorithms like FLOE may outperform them. Importantly, the UP betting strategy is invariant to the order of observations $Z_1,...,Z_t$ \citep{orabona2022tightconcentrationsconfidencesequences}, which suggests it weights $V_1$ and $V_t$ identically. However, as mentioned in our motivation for FLOE, we should avoid this; late payoffs are based on a far better understanding of the distribution than early ones.

\begin{figure}[!hbt]
  \centering
    \begin{subfigure}{\linewidth}
      \centering
      \includegraphics[width=0.7\linewidth]{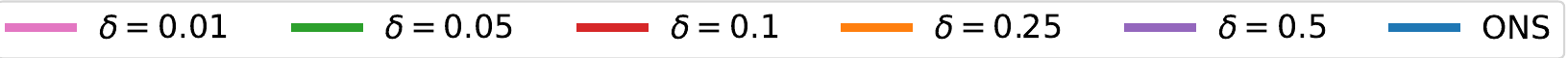}
    \end{subfigure}
  \begin{subfigure}[b]{\linewidth}
    \centering
    \includegraphics[width=\linewidth]{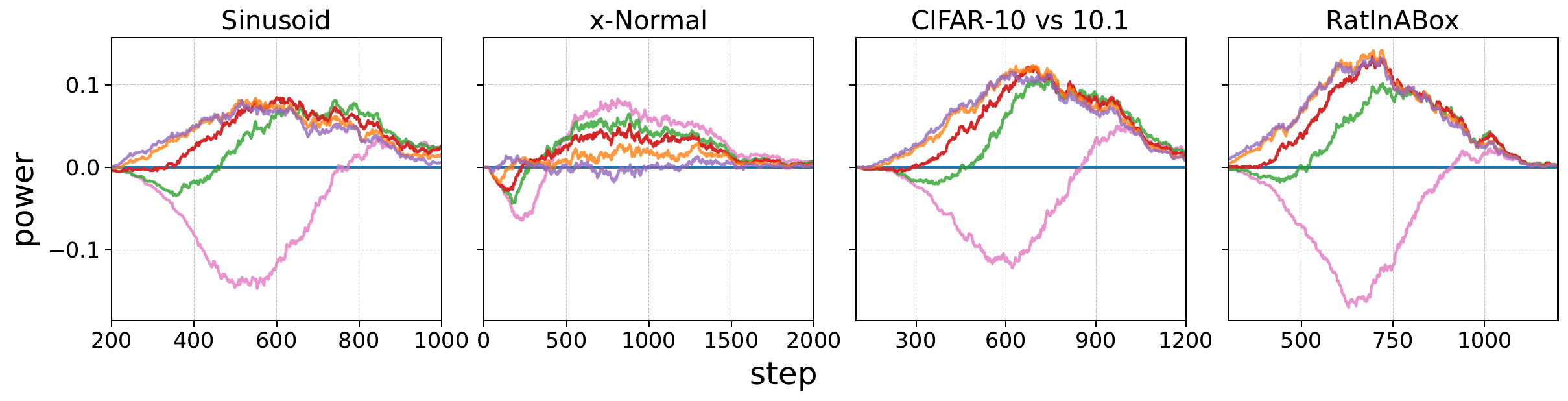}
  \end{subfigure}
  \caption{Power of FLOE-R relative to ONS at varying regularization levels $\delta$ and fixed $\alpha=0.1$.}
  \label{fig:sensitivity_delta_rel_power}
  \vspace{-1em}
\end{figure}

\textbf{Hyperparameter Sensitivity.}
We would like to confirm that our regularizer is robust to the choice of hyperparameters $\alpha,\delta_0,\delta_1 \in (0,1)$. To compartmentalize the ablation we fix $\alpha=0.1$ and vary $\delta:=\delta_0 = \delta_1$ between two extremes: $0.01$ and $0.5$. \cref{fig:sensitivity_delta_rel_power} illustrates the difference in power between FLOE at each level $\delta$ and ONS. In all problems only the extreme choice $\delta=0.01$ results in lower power than ONS. 
The other regularization hyperparameter to consider is $\alpha$. For this ablation we fix $\delta_0 = \delta_1 = 0.1$ and vary $\alpha$ between $0.05$ and $0.5$; these results are shown in \cref{fig:sensitivity_alpha_rel_power} of the appendix.
Overall, we find that our choice of parameters $\alpha = \delta_0 = \delta_1 = 0.1$ is a good middle ground for regularization strength, and that parameter choices around $0.1$ and above all tend to be fairly robust in terms of performance.
When $\alpha$ and $\delta$ are very large (e.g. $>0.5$), however, the regularizer impact is minimal.

\textbf{FLOE for Sequential Kernel Independence Testing.}
Nothing about FLOE is specific to two-sample testing. The generality of the leave-one-out approach can easily be applied to learning critic functions for other problems, such as independence testing. We illustrate this by applying FLOE to the sequential kernel independence test (SKIT) of \citep{podkopaev2023seqhsic}. Similar to the kernel two-sample case, the leave-one-out payoffs for SKIT can be computed in cumulative quadratic time and memory. Exact expressions and implementation details are provided in \cref{app:floe-skit}. 

\begin{figure}[hbt!]
  \centering
  \begin{subfigure}{\linewidth}
    \centering
    \includegraphics[width=0.5\linewidth]{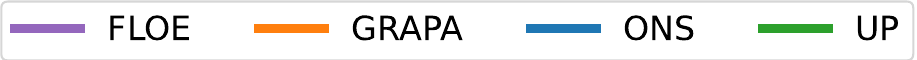}
  \end{subfigure}
  \begin{subfigure}[b]{\linewidth}
    \centering
    \includegraphics[width=0.7\linewidth,trim={0 2cm 0 0},clip]{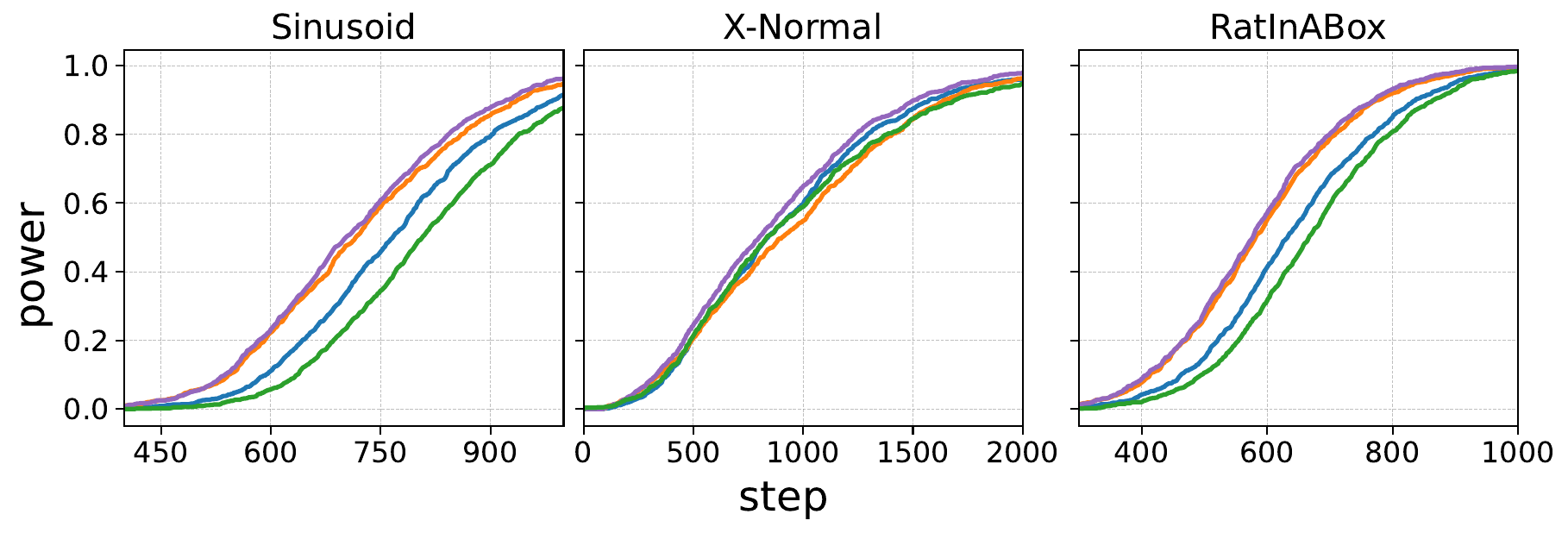}
  \end{subfigure}
  \begin{subfigure}[b]{\linewidth}
    \centering
    \includegraphics[width=0.7\linewidth,trim={0 0 0 9mm},clip]{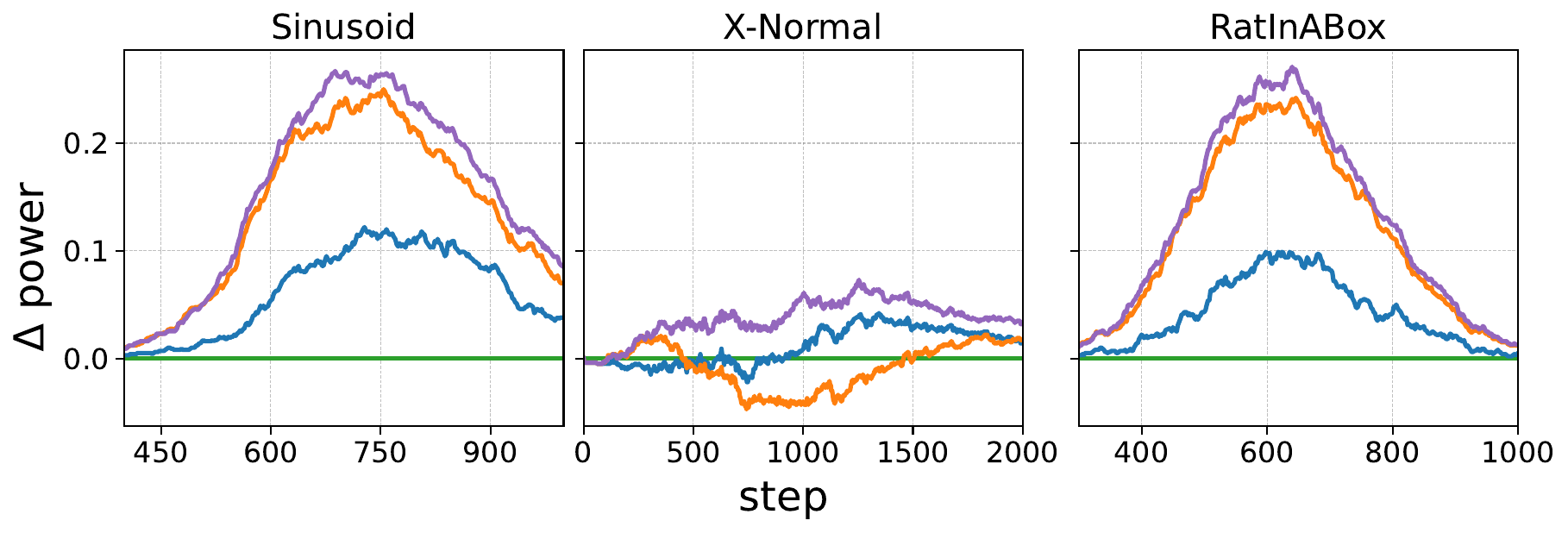}
  \end{subfigure}
  \caption{Mean test power on a variety of datasets; each test is run 1,000 times. The second row subtracts the performance of UP from each point for easier visualization.}
  \label{fig:baselines_floe_skit}
\end{figure}

We evaluate FLOE on independence-testing versions of the Sinusoid, x-Normal, and RatInABox problems. For the former two, we sample $X$ and $Y$ jointly from the 2D distribution containing component-wise dependency. For RatInABox, we compare the firing rates of 6 place cells against 6 boundary vector cells. Detailed problem and test settings are given in \cref{app:floe-skit-exps}. The performance of FLOE against GRAPA, ONS, and Universal Portfolio with a Beta($1/2,1/2$) prior is plotted in \cref{fig:baselines_floe_skit}. Results show FLOE outperforms these other baselines even without the use of regularization.

\section{Conclusion}
\label{sec:conclusions}
Existing betting strategies in sequential anytime-valid testing default to a handful of classical online learning algorithms with nice theoretical adversarial-regime guarantees. In the stochastic setting, however, standard FTL algorithms may perform better in practice. Therefore,
we framed the problem of selecting betting strategies in online testing
in such a way as to very naturally motivate following a better estimate of the e-power.
Our proposed method, FLOE, is natural, simple to implement,
computationally efficient for kernel tests,
and leads to substantial improvements in test power over prior work
at very little computational cost.

\begin{ack}
The authors would like to thank Zheng He for productive discussions,
and the anonymous reviewers for insights that improved the paper.

This work was enabled in part by the Natural Sciences and Engineering Research Council of Canada,
the Canada CIFAR AI Chairs program,
the Digital Research Alliance of Canada,
and Calcul Québec.
Feng Liu is supported by the ARC with grant numbers DP230101540, DE240101089 and DP260103996.
\end{ack}

\printbibliography

\appendix
\counterwithin*{figure}{section}
\renewcommand{\thefigure}{\thesection.\arabic{figure}}
\renewcommand{\theHfigure}{\thesection.\arabic{figure}}

\newpage
\section{Proofs}

\begin{lemma}[Indirect E-Power Regularization]
    \label[lemma]{lem:indirect-epower-reg}
    Any $(\alpha,\beta)$-regularized set of bets $\Lambda^{(\alpha,\beta)} \subseteq [0,1-\epsilon]$ for some $\epsilon\in(0,1)$ also regularizes the e-power according to 
    \begin{align*}
        \Pr_{\bZ_{t-1}} \left( \inf_{\lambda\in\Lambda^{(\alpha,\beta)}} \bE[\log(1 + \lambda V_t ) \mid \sF_{t-1}]  > \log(1 - \alpha) - \frac{1}{2\epsilon^2} \Var[V_t \mid \sF_{t-1}]  \right) \geq 1 - \beta.
    \end{align*}
\end{lemma}

\begin{proof}
    The idea is to apply Taylor's theorem with a second-order Lagrange remainder term to $\log(1 + \lambda V_t)$ about $\bE[\lambda V_t \mid \sF_{t-1}]$. For brevity, we will assume all expectations are conditioned on the filtration $\sF_{t-1}$, even if not explicitly written. 

    The Taylor expanded expression for the e-power is
    \begin{align*}
        \bE[\log(1 + \lambda V_t)] = \log(1 + \bE[\lambda V_t]) - \bE\left[ \frac{(\lambda V_t - \bE[\lambda V_t])^2}{2(1 + \zeta)^2} \right],
    \end{align*}
    for some random variable $\zeta$ between $\lambda V_t$ and $\bE[\lambda V_t]$. Since $|\lambda V_t| \le 1 - \epsilon$, it follows that
    \begin{align*}
        \bE[\log(1 + \lambda V_t)] \ge \log(1 + \bE[\lambda V_t]) - \frac{\Var[V_t]}{2\epsilon^2}.
    \end{align*}
    Finally, taking the infimum over the regularization set $\lambda \in \Lambda^{(\alpha,\beta)}$ and applying a union bound with $\Pr( \inf_{\lambda \in \Lambda^{(\alpha, \beta)}} \bE[\lambda V_t] > -\alpha) \ge 1 - \beta$ yields, with probability $1-\beta$,
    \begin{align*}
        \inf_{\lambda\in\Lambda^{(\alpha,\beta)}} \bE[\log(1 + \lambda V_t)] \ge \log(1 - \alpha) - \frac{\Var[V_t]}{2\epsilon^2}
    ,\end{align*}
    as desired.
\end{proof}

\subsection{Test Consistency}
\label[appendix]{app:test_consistency}

\begin{lemma}
    \label[lemma]{lem:loo-conv}
    Under the alternative hypothesis with a characteristic kernel, the leave-one-out predictor $\nu_\ell^{(t)} := \nu_{[t]\setminus\{\ell\}}$ defined in \cref{eq:loo_predictor} converges uniformly to the witness predictor $\nu^*$ defined in \cref{eq:witness_predictor} for any left-out sample $\ell\le t$.
    \[
        \sup_{z\in\cZ} \left| \nu_\ell^{(t)}(z) - \nu^*(z) \right| \xrightarrow{a.s.} 0
        \qquad
        \forall \ell \le t.
    \]
\end{lemma}
\begin{proof}
    Since the RKHS $\cF$ is a strong Glivenko-Cantelli class \citep{gretton12a_mmd, bartlett_rademacher}, so too is $\cH = \{h(z, \cdot): z\in\cZ\}$ as it is a linear composition of $\cF$. This means $\cH$ satisfies the uniform law of large numbers
    \[
        \sup_{z\in\cZ} \left| \frac{1}{t} \sum_{i=1}^t h(z, Z_i) - \bE[h(z,Z)] \right|
        \xrightarrow{a.s.} 0
    \]
    If we let $g_t:=\frac{1}{t}\sum_{i=1}^t h(\cdot, Z_i) $ and $ g:=\bE_{Z}[h(\cdot,Z)] $, this is equivalent to saying $ \| g_t - g \|_\infty \xrightarrow{a.s.} 0 $. %

    Let $\eta_t := \frac{1}{t^2}\sum_{i=1}^t \sum_{j=1}^t h(Z_i,Z_j)$ and $\eta:=\bE[h(Z,Z')]$. The strong law of large numbers for U-Statistics \citep{hoeffding1961_strong} tells us $| \eta_t - \eta | \xrightarrow{a.s.} 0 $. Also note that $\eta$ is just the population MMD which, under the alternative hypothesis with a characteristic kernel, is positive. 

    It follows that $(g_t,\eta_t) \xrightarrow{a.s.} (g, \eta)$ in the product space, and so applying the continuous mapping theorem with $\Phi = (g,\eta) \mapsto g/\sqrt{2K\eta}$ yields
    \[
        \left\| \frac{g_t}{\sqrt{2K\eta_t}} - \frac{g}{\sqrt{2K\eta}} \right\|_\infty \xrightarrow{a.s.} 0.
    \]
    Notice that $\nu^* = g/\sqrt{2K\eta}$ and $\nu_\ell^{(t)} = g_t / \sqrt{2K\eta_t} + o(1)$ for any $\ell\le t$, and so $\max_{\ell\le t} \| \nu_\ell^{(t)} - \nu^* \|_\infty \xrightarrow{a.s.} 0 $.
\end{proof}

\begin{proposition}
    \label[proposition]{prop:loo-epower-conv}
    Let $V_\ell^{(t)}$ be the loo payoffs and $f^*$ be the witness function of \cref{eq:witness_rkhs} with a characteristic kernel. Also,
    suppose the bets $\lambda$ are in a bounded set $[0,1-\epsilon]$ for some $\epsilon > 0$. Then, under the alternative hypothesis, we have
    \[
        \sup_{\lambda \in [0,1-\epsilon]} \left| \frac{1}{t} \sum_{\ell=1}^t \log( 1 + \lambda V_\ell^{(t)} ) - \bE\left[ \log\left(1 + \lambda \left(f^*(X) - f^*(Y) \right) \right) \right] \right| 
        \xrightarrow{a.s.} 
        0
    \]
\end{proposition}

\begin{proof} 
    Let $\psi_\lambda(V) := \log(1 + \lambda V)$ and recall that we defined $V_\ell^* := \nu^*(Z_\ell)$. Since $\lambda < 1$ and $V_\ell^{(t)} \in [-1,1]$ the function $\psi_\lambda$ is $L$-Lipschitz for some $L > 0$. Define the sup-norm $\|\psi_\lambda(V)\|_\infty = \sup_\lambda | \psi_\lambda(V) |$ with respect to $\lambda$, and apply the decomposition    
    \begin{align*}
        \left\| \frac{1}{t} \sum_{\ell=1}^t  \psi_\lambda(V_\ell^{(t)}) - \bE[ \psi_\lambda(V_1^*) ] \right\|_\infty
        \le \left\| \frac{1}{t} \sum_{\ell=1}^t \psi_\lambda(V_\ell^{(t)}) - \psi_\lambda(V_\ell^*) \right\|_\infty + 
        \left\| \frac{1}{t} \sum_{\ell=1}^t \psi_\lambda(V_\ell^*) - \bE[\psi_\lambda(V_1^*)] \right\|_\infty.
    \end{align*}
    For the first term, we can use the Lipschitz property to write
    \begin{align*}
        \left\| \frac{1}{t} \sum_{\ell=1}^t \psi_\lambda( V_\ell^{(t)} ) - \psi_\lambda(V_\ell^{*}) \right\|_\infty
        &\le 
        \frac{1}{t} \sum_{\ell=1}^t \left\| \psi_\lambda(V_\ell^{(t)}) - \psi_\lambda(V_\ell^{*}) \right\|_\infty 
        \le \frac{1}{t} \sum_{\ell=1}^t \sup_\lambda \left| \psi_\lambda(V_\ell^{(t)}) - \psi_\lambda(V_\ell^*) \right| \\
        &\le \frac{1}{t} \sum_{\ell=1}^t L \left| V_\ell^{(t)} - V_\ell^{*} \right| \\
        &\le L \max_{i\le t} \left| V_i^{(t)} - V_i^{*} \right| \\
        &\le L \max_{i\le t} \sup_{z\in\cZ} \left| \nu_i^{(t)}(z) - \nu^*(z) \right|.
    \end{align*}
    By \cref{lem:loo-conv} this must also converge a.s. to zero.
    The second term converges a.s. to zero by the uniform law of large numbers since $\{\psi_\lambda: \lambda \in [0,1-\epsilon]\}$ is a strong Glivenko-Cantelli class when the domain is bounded in $[-1,1]$. This is easily verified by noting that the Rademacher complexity of this function class is sub-linear.
\end{proof}

\maintheorem*

\begin{proof}
    It suffices to show that the e-power converges to some positive value, i.e.
    \[
        \lim_{t\to\infty} \bE[\log(1 + \lambda_t V_t^{(t)} ) \mid \sF_{t-1}] \overset{a.s.}{>} 0,
    \]
    then consistency of the test directly follows Proposition 4.2 of \citet{pandeva24_deep}.

    Since the bets are bounded away from 1 in an open interval, they must be a subset of $[0,1-\epsilon]$ for some $\epsilon\in(0,1)$.
    Note that $\lambda_t$ exists and is unique on the set $[0,1-\epsilon]$ by concavity of the log function. Thus, it follows from the uniform convergence result of \cref{prop:loo-epower-conv} that
    \begin{align*}
        \lambda_t \xrightarrow{a.s.} \lambda^* := \argmax_{\lambda\in[0,1-\epsilon]} \bE[ \log(1 + \lambda ( f^*(X) - f^*(Y) ) ) ].
    \end{align*}
    Since the data is i.i.d., the conditional e-power is equivalent to
    $\bE_Z[\log(1 + \lambda_t \cdot v_t^{(t)}(Z))] $. We know from above that $\lambda_t \xrightarrow{a.s.} \lambda^*$, and \cref{lem:loo-conv} tells us $v_t^{(t)}(Z) \xrightarrow{a.s.} v^*(Z)$. As such, applying the continuous mapping theorem and dominated convergence theorem tells us
    \[
        \bE[\log(1 + \lambda_t V_t^{(t)}) \mid \sF_{t-1}] \xrightarrow{a.s.} \bE[ \log(1 + \lambda^* ( f^*(X) - f^*(Y) ) ) ].
    \]
    This limit is just $\max_{\lambda\in[0,1-\epsilon]} \bE[ \log(1 + \lambda ( f^*(X) - f^*(Y) )) ] $. Taking the Taylor expansion of $\log(1+\lambda V)$ about 0 gives
    \[
        \lambda \bE[V^*] - \frac{\lambda^2 \bE[(V^*)^2]}{2!} + \frac{\lambda^3 \bE[(V^*)^3]}{3!} - \cdots 
    \]
    which, under the alternative hypothesis, is positive for some small enough $\lambda > 0$ since $\bE[V^*] = \MMD(\bP,\bQ) > 0$.
\end{proof}

\newpage
\subsection{Bet Regularization}
\label[appendix]{app:bet_reg}

\begin{lemma}[Bennett tail bound]
    \label[lemma]{lem:bennett}
    If $X - \bE[X] \le b$ and $\sigma^2 := \Var[X] < \infty$ then
    \[
        \bE[e^{\lambda (X - \bE[X])}] \le \exp(\lambda^2 \sigma^2 g(\lambda b))
        \quad
        \forall \lambda > 0,
    \]
    where $g = u \mapsto (e^u - u -1) / u^2$.
\end{lemma}

\begin{proof}
    Take the Taylor expansion of $e^{\lambda X}$ and use the fact $g(u) = \sum_{k=2}^\infty \frac{u^{k-2}}{k!}$ 
    \begin{align*}
        \bE[ e^{\lambda (X - \bE[X])} ]
        &\le 1 + \bE[ \lambda^2 (X-\bE[X])^2 g(\lambda b) ]
        \le 1 + \lambda^2 \sigma^2 g(\lambda b).
    \end{align*}
    The result follows because $1+x\le e^x$.
\end{proof}

\begin{proposition}
    \label[proposition]{prop:epower-bound}
    Let $\Tilde{V}_{t+1} = \frac{1}{t} \sum_{i=1}^t h(Z_i, Z_{t+1}) $ be the unnormalized ERM-payoff at step $t+1$ for i.i.d. data $Z_1,Z_2,...$, and let $\sF_t = \sigma(\{Z_1,...,Z_{t}\})$. Then for any error rate $\delta\in(0,1)$
    \begin{align*}
        \Pr\left( \bE[\Tilde{V}_{t+1} \mid \sF_{t}] > \MMD^2(\bP,\bQ) - \frac{\zeta}{4} \phi_+^{-1}\left( \frac{16 \log 1/\delta}{t\zeta }  \right) \right) \ge 1 - \delta,
    \end{align*}
    where $\phi_+^{-1}$ is the positive inverse of $\phi(u) = (1+u)\log(1+u) - u$, and $\zeta = \Var_{Z}[ \bE_{\Tilde{Z}}[ h(Z,\Tilde{Z})]]$.
\end{proposition}

\begin{proof}
    Let $U_i := \bE[ \Tilde{V}_{t+1} \mid \sF_{i} ] $ and note the sequence $(U_0, U_1,...,U_t)$ forms a Doob martingale. First, the Martingale differences $U_i - U_{i-1}$ are bounded below by $-4/t$ since
    \begin{align*}
        U_i - U_{i-1}
        &= \frac{1}{t} \left( \bE[h(Z_i,Z_{t+1}) \mid \sF_i] - \bE[ h(Z_i, Z_{t+1}) \mid \sF_{i-1} ] \right) \\
        &= \frac{1}{t} \left( \bE[h(z_i, Z)] - \bE[h(Z,Z')] \right) \ge -\frac{4}{t}.
    \end{align*}
    Also, its conditional variance is
    \[
        \Var[U_i - U_{i-1} \mid \sF_{i-1}] = \frac{1}{t^2} \Var_Z[\bE_{\Tilde{Z}}[h(Z,\Tilde{Z})]] = \frac{\zeta}{t^2}.
    \]
    Hence the conditional differences satisfy a Bennett tail bound, a fact we now use to derive the result.
    
    For any $\epsilon\in\bR$, we have the following Chernoff bound on the lower tail:
    \begin{align*}
        \Pr( U_0 - U_t \ge \epsilon )
        \le \inf_{\lambda>0} e^{-\lambda \epsilon} \bE[e^{\lambda (U_0 - U_t) }].
    \end{align*}
    The expectation in the bound can be unrolled into a composition of martingale differences, each of which is bounded using \cref{lem:bennett} with $b:=4/t$ and $\sigma^2 := \zeta/t^2$ to produce
    \begin{align*}
        \bE[ e^{\lambda(U_0 - U_t)} ]
        &= \bE[ e^{\lambda \sum_{i=1}^t U_{i-1} - U_i } ] \\
        &= \bE[ e^{\lambda \sum_{i=1}^{t-1} U_{i-1} - U_{i} } \bE[ e^{\lambda (U_{t-1} - U_t)} \mid \sF_{t-1} ] ] \\
        &\le \exp\left( \lambda^2 t \sigma^2  g(\lambda b) \right).
    \end{align*}
    Plugging this back into our Chernoff bound yields
    \begin{align*}
        \Pr( U_0 - U_t \ge \epsilon) 
        &\le \inf_{\lambda>0}  \exp(-\lambda\epsilon + \lambda^2 t\sigma^2 g(\lambda b))
        = \exp\left( -\frac{t\sigma^2}{b^2} \phi\left( \frac{b\epsilon}{t\sigma^2} \right) \right),
    \end{align*}
    where the minimizer can be derived by noticing that the expression in the exponent is convex in $\lambda$; it is $\lambda^* = \frac{1}{b}\log(1 + \frac{b\epsilon}{t\sigma^2})$. This satisfies $\lambda^* > 0$ for any $\epsilon > 0$. Finally, inverting this tail bound, noticing $U_0 = \MMD^2(\bP,\bQ)$ and substituting in the values for $b$ and $\sigma^2$ gives us the desired result.
\end{proof}

\theorembet*

\begin{proof}
    Let $\Lambda(\bZ_t)$ be the proposed regularized bet set. We want to show that with confidence $1 - (\delta_0+\delta_1+\delta_2)$
    \[
        \inf_{\lambda \in \Lambda(\bZ_t)}  \bE[ \alpha + \lambda V_{t+1} \mid \sF_t ] > 0.
    \]
    To make things more readable, we omit dependencies on $\bP\times\bQ$ and $\bZ_t$, and we let $\eta_t := \hMMD[2]_b = \frac{1}{t^2} \sum_{i=1}^t \sum_{j=1}^t h(Z_i,Z_j)$ be the biased MMD estimator.
    
    Notice that for all $\lambda\in \Lambda(\bZ_t)$
    \[
        -\frac{\alpha\sqrt{2K\eta_t}}{\lambda} < \tMMD[2]_b - \frac{\widetilde{\zeta}_b}{4} \phi_+^{-1}\left( \frac{16\log1/\delta_0}{\widetilde{\zeta}_b t} \right),
    \]
    which is easily verified by solving this bound for $\lambda$. The term on the right is is increasing with $\tMMD[2]_b$ and decreasing with $\widetilde{\zeta}_b$, so by taking a union bound with the events $\{ \tMMD[2]_b \le \MMD^2 \}$ and $\{ \widetilde{\zeta}_b \ge \zeta \}$ we get with confidence $1-(\delta_1+\delta_2)$ that
    \[
        -\frac{\alpha \sqrt{2K\eta_t}}{\lambda} < \MMD^2 - \frac{\zeta}{4} \phi_+^{-1}\left( \frac{16\log1/\delta_0}{\zeta t} \right)
        \quad
        \forall \lambda\in\Lambda(\bZ_t).
    \]
    By \cref{prop:epower-bound} the term on the right is upper bounded by $\bE[\Tilde{V}_{t+1}\mid \sF_t]$ with probability $1-\delta_0$, and so a union bound with this event gives us with confidence $1-(\delta_0+\delta_1+\delta_2)$ that
    \[
        -\frac{\alpha\sqrt{2K\eta_t}}{\lambda} < \bE[\Tilde{V}_{t+1} \mid \sF_t]
        \quad
        \forall \lambda\in\Lambda(\bZ_t).
    \]
    Rearranging terms gives us the desired result.
\end{proof}

\newpage

\section{FLOE for Sequential Kernel Independence Testing} \label{app:floe-skit}
We give an overview of the sequential kernel independence test, and refer readers to \citep{podkopaev2023seqhsic} for a more detailed explanation.

\textbf{Oracle Critic.}
Similar to the kernel two-sample test, the oracle function for the independence test is the supremum over bounded Hilbert-Schmidt operators from RKHS $\cH_X$ to RKHS $\cH_Y$. 

Let $k$ and $l$ be the kernels associated with $\cH_X$ and $\cH_Y$, respectively, and let $\phi(X) := k(\cdot,X)$ and $\varphi(Y) := l(\cdot, Y)$ be their canonical feature maps. Also suppose $\sup_{x,x'}k(x,x') \le K_X$ and $\sup_{y,y'}l(y,y') \le K_Y$.
The Hilbert-Schmidt Independence Criterion (HSIC) \citep{gretton_hsic} is

\begin{align*}
    \sqrt{\text{HSIC}(X,Y)} = \sup_{\|F\| \le 1} \inner{F}{C_{YX}}_{\text{HS}(\cH_X,\cH_Y)}
\end{align*}
where $C_{YX} = \bE[ \varphi(Y) \otimes \phi(X) ]$ is the covariance operator defined with $[\varphi(Y) \otimes \phi(X)](f) = \inner{f}{\phi(X)} \varphi(Y)$. Thus, the oracle critic is the normalized covariance operator $C_{YX} / \|C_{YX}\|$.

Suppose we have $2n$ i.i.d. samples $(X_1,Y_1),...,(X_{2n}, Y_{2n})$. For convenience, we rename the latter $n$ samples by $(X_1',...,X_n')$ so that we now have two sets of samples $(X_i,Y_i)_{i=1..n}$ and $(X_i',Y_i')_{i=1..n}$. 

A test martingale constructed with this oracle critic has payoffs given by
\begin{align*}
    V_t = \frac{1}{2K_XK_Y}\inner{\frac{C_{YX}}{\|C_{YX}\|}}{ ( \varphi(Y_t) - \varphi(Y_t') ) \otimes ( \phi(X_t) - \phi(X_t') ) },
\end{align*}
which is easily verified by checking $|V_t| \le 1$ and $\bE_{\fH_0}[V_t \mid \sF_{t-1}] = 0$.

\textbf{Loo Payoffs.}
Let $\hat{C}_{YX}^{[t]\backslash\{\ell\}}$ be an estimate of the covariance operator from all samples indexed by $1,...,t$ excluding the samples $(X_\ell, Y_\ell)$ and $(X_\ell',Y_\ell')$. For brevity, we will denote $\hat{\bE}_{i\ne\ell}^{t}[a_i] = \frac{1}{2(t-1)} \sum_{i=1; i\ne \ell}^t a_i$. Then,
\begin{align*}
    \hat{C}_{YX}^{[t]\backslash\{\ell\}} = \,\,&\hat{\bE}_{i\ne\ell}^t[ \varphi(Y_i) \otimes \phi(X_i) + \varphi(Y_i') \otimes \phi(X_i')] \\
    &- \hat{\bE}_{k\ne\ell}^t[\varphi(Y_k) + \varphi(Y_k')] \otimes \hat{\bE}_{j\ne\ell}^t[ \phi(X_j) + \phi(X_j') ].
\end{align*}

It follows that the loo-payoffs $V_\ell^{(t)}$, which is the loo-critic $\hat{C}_{YX}^{[t]\backslash\{\ell\}}$ evaluated at its held-out samples $(X_\ell,Y_\ell)$ and $(X_\ell',Y_\ell')$, is
\begin{align*}
    V_\ell^{(t)} &= \frac{1}{2K_XK_Y \|\hat{C}_{YX}^{[t]\backslash\{\ell\}}\| } \inner{ \hat{C}_{YX}^{[t]\backslash\{\ell\}} }{ ( \varphi(Y_\ell) - \varphi(Y_\ell') ) \otimes ( \phi(X_\ell) - \phi(X_\ell') ) } \\
    &= \frac{1}{2K_XK_Y \|\hat{C}_{YX}^{[t]\backslash\{\ell\}}\| }
    \Bigg\{
        \hat{\bE}_{i\ne\ell}^t\left[ \big( k(X_i,X_\ell) - k(X_i,X_\ell') \big)\big( l(Y_i,Y_\ell) - l(Y_i,Y_\ell') \big) \right] \\
        &\qquad\qquad\qquad\qquad\qquad
        + \hat{\bE}_{i\ne\ell}^t \left[ \big( k(X_i',X_\ell) - k(X_i',X_\ell') \big) \big( l(Y_i',Y_\ell) - l(Y_i',Y_\ell') \big) \right] \\
        &\qquad\qquad\qquad\qquad\qquad
        - \hat{\bE}_{i\ne\ell}^t \left[ \big( k(X_i,X_\ell) - k(X_i,X_\ell') + k(X_i',X_\ell) - k(X_i',X_\ell') \big) \right] \\
        &\qquad\qquad\qquad\qquad\qquad
        \times \hat{\bE}_{i\ne\ell}^t \left[ \big( l(Y_i,Y_\ell) - l(Y_i,Y_\ell') + l(Y_i',Y_\ell) - l(Y_i',Y_\ell') \big) \right]
    \Bigg\}.
\end{align*}
Note that $\|\hat{C}_{YX}^{[t]\backslash\{\ell\}}\|$ is exactly the root biased HSIC estimator from samples $1..t$ excluding $\ell$.

\textbf{Implementation Details.}
Similarly to the kernel two-sample test of \cref{sec:pratical_test}, this independence test can be done in linear time per step as long as we store the kernel matrices $K_{ij} = k(X_i,X_j)$ and $L_{ij} = l(Y_i,Y_j)$ of the previous step. Cumulatively, the test is quadratic in both compute and memory. Maintaining the latter is possible by subtracting the sums where the left-out index $\ell$ is included. For example, in $\| \hat{C}_{YX}^{[t]\backslash\{\ell\}  }\|$ we can simply compute the HSIC estimate and then subtract the terms that include index $\ell$ in them. Since HSIC is composed of nested sums, the principle of inclusion-exclusion is used.

\subsection{Experiments}
\label{app:floe-skit-exps}

We compare the performance of FLOE against GRAPA \citep{waudbysmith2022estimatingmeansboundedrandom}, ONS \citep{hazan2007ons}, and universal portfolio with Beta(1/2,1/2) prior (UP) \citep{cover1991, orabona2022tightconcentrationsconfidencesequences} betting strategies, and report the results in \cref{fig:baselines_floe_skit}. We use Gaussian kernels for both $\cH_X$ and $\cH_Y$ and select the bandwidths to optimize the performance of ONS. We find FLOE performs the best on our considered datasets.

\textbf{Sinusoid} \citep{sejdinovic2012hypothesistestingusingpairwise}\textbf{.}
We test for dependencies between $X$ and $Y$ sampled jointly from $\propto 1 + \sin(\omega x) \sin(\omega y)$ on the support $[-\pi,\pi]^2$. Higher frequencies $\omega$ produce subtler dependencies, resulting in a harder independence test.
We use $\omega=5$.

\textbf{x-Normal.}
We jointly sample $X$ and $Y$ from the xNormal distribution used in our two-sample testing experiments. Smaller correlation coefficients result in harder-to-detect dependencies. We use a coefficient of $\pm0.5$ for its normal components.

\textbf{RatInABox} \citep{ratinabox}\textbf{.}
We compare the firing rates of place cells (neurons that are activated in the vicinity of certain locations) to those of boundary vector cells (neurons that fire when the rat is at a specific distance from a wall). These cell activation locations are not disjoint, and so there should be dependencies between their firing rates. We use a standard 1x1 box environment with a protruding wall, and sample from 6 place cells and 6 boundary vector cells.

\section{Additional Experiments and Details} \label{app:more-experiments}

\begin{figure}[hbt!]
  \centering
  \begin{subfigure}{\linewidth}
    \centering
    \includegraphics[width=0.7\linewidth]{figures/benchmark/floe/two-sample/legend.pdf}
  \end{subfigure}
  \begin{subfigure}[t]{0.25\linewidth}
    \centering
    \includegraphics[width=\linewidth]{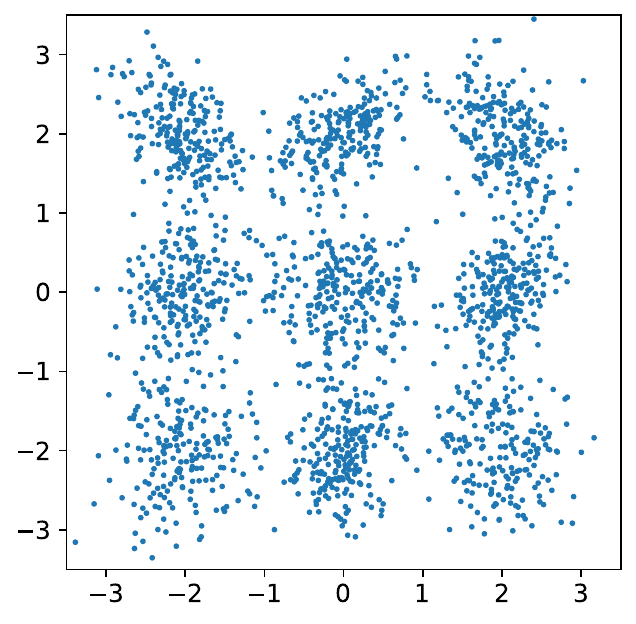}
  \end{subfigure}
  \begin{subfigure}[t]{0.6\linewidth}
    \centering
    \includegraphics[width=\linewidth]{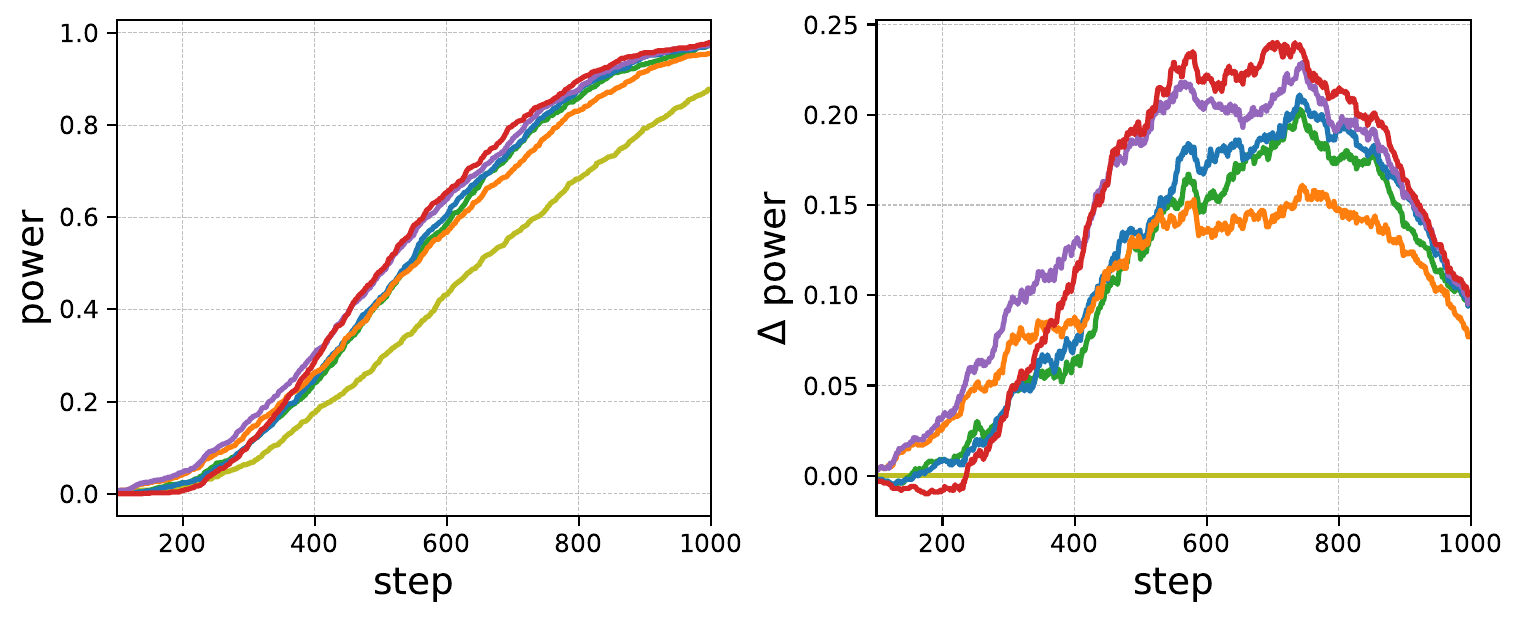}
  \end{subfigure}
  \caption{Blob dataset: On the left are samples from $\bQ$. In the middle is the mean power at time $t$. On the right is the power relative to SR.}
  \label{fig:baselines_blobs}
\end{figure}

\textbf{Blobs.}
We perform an additional evaluation on the Blobs dataset \citep{chwialkowski2015tst, gretton2012optkernal}, which is a benchmark synthetic problem composed of a 3x3 uniform grid of bivariate Gaussians. These Gaussian `blobs' are isotropic in $\bP$ and anisotropic in $\bQ$ with each blob randomly assigned an eigenvalue ratio and rotation angle. The difficulty lies in the fact the discrepancy between distributions are encoded at a much smaller frequency than the overall data. Results are plotted in \cref{fig:baselines_blobs} for the average power over 1,000 tests. FLOE-R performs better than FLOE at moderate to large sample sizes, and both are better than ONS and GRAPA.

\textbf{Higgs} \citep{baldi2014higgs}\textbf{.}
We aim to identify if a Higgs boson has been created by comparing the signals of processes that produce it to background processes that do not. In particular, we focus on the $\varphi$-momenta of four particle jets, which hold very little discriminating information for whether a Higgs boson was birthed.

\begin{figure}[hbt!]
  \centering
  \begin{subfigure}{\linewidth}
    \centering
    \includegraphics[width=0.7\linewidth]{figures/benchmark/floe/two-sample/legend.pdf}
  \end{subfigure}
  \begin{subfigure}[t]{0.6\linewidth}
    \centering
    \includegraphics[width=\linewidth]{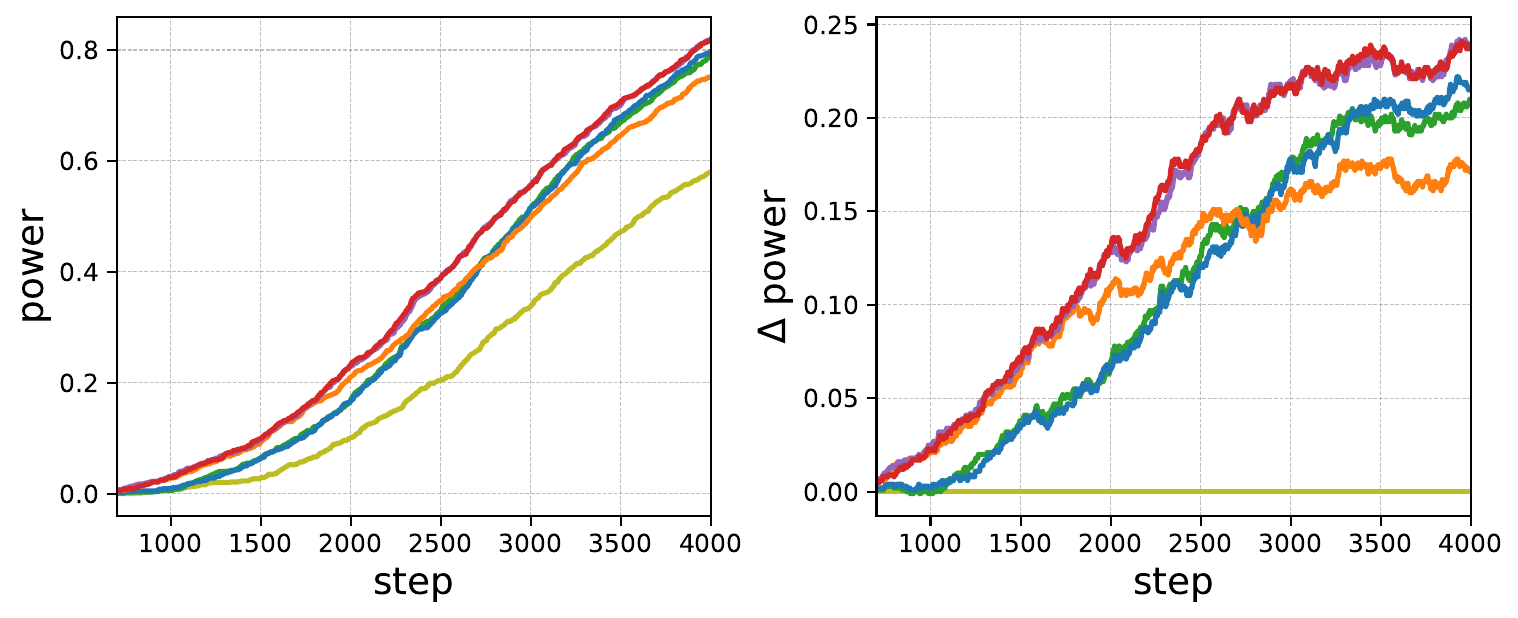}
  \end{subfigure}
  \caption{Higgs dataset. The plot on the right subtracts the power of SR from each baseline for easier visualization.}
  \label{fig:baselines_higgs}
\end{figure}

\begin{figure}[hbt!]
  \centering
  \begin{subfigure}[t]{0.4\linewidth}
    \centering
    \includegraphics[width=\linewidth]{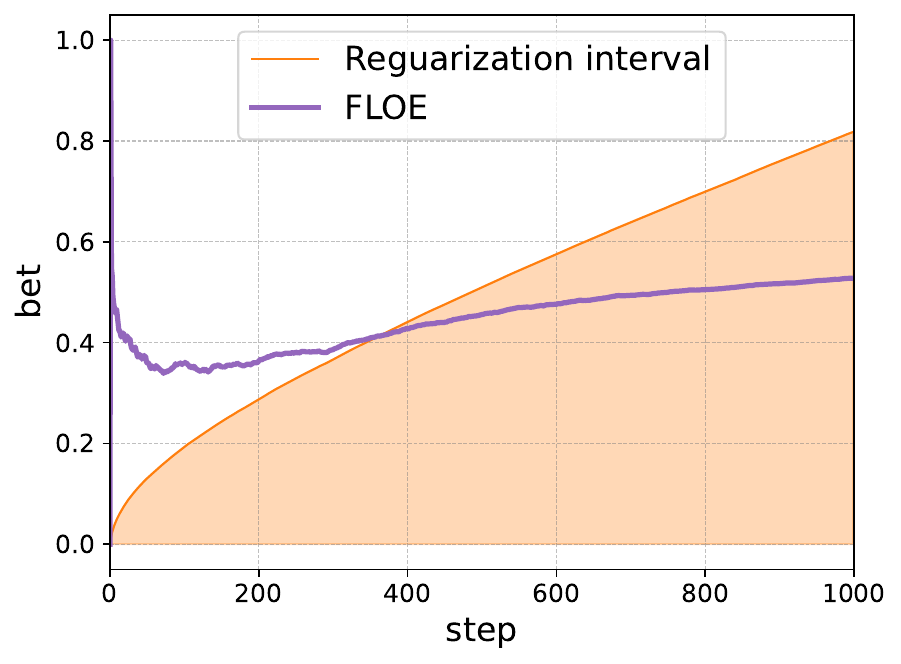}
    \caption{The average regularization interval used by FLOE-R for the xNormal dataset.}
    \label{fig:bet_interval}
  \end{subfigure}
  \begin{subfigure}[t]{0.25\linewidth}
    \centering
    \includegraphics[width=\linewidth]{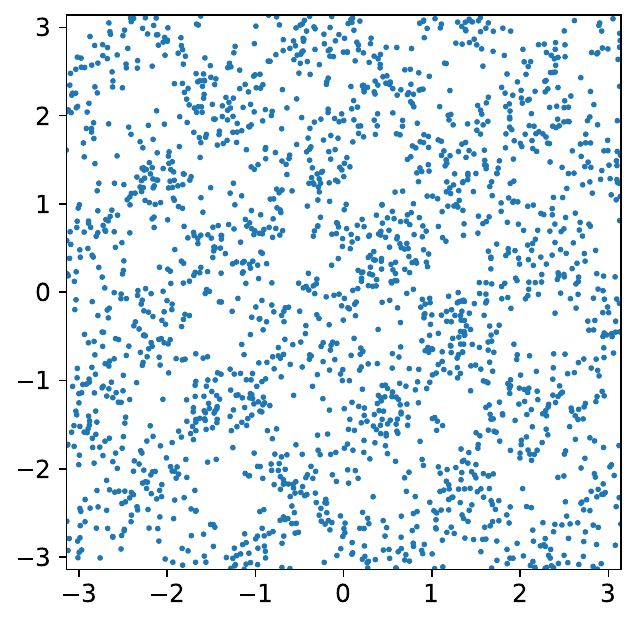}
    \caption{Sinusoid}
    \label{fig:density_sinusoid}
  \end{subfigure}
  \begin{subfigure}[t]{0.25\linewidth}
    \centering
    \includegraphics[width=\linewidth]{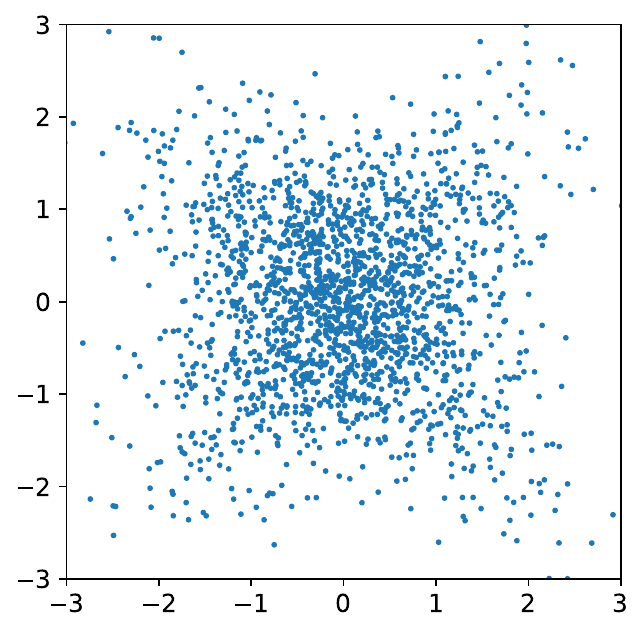}
    \caption{X-Normal}
    \label{fig:density_hdgm}
  \end{subfigure}
  \caption{(a) Visualization of FLOE's bet value and the regularization interval used for FLOE-R. (b)-(c) Samples from the toy datasets.}
\end{figure}

\textbf{Visualized Bet Interval.}
It will be informative to see how the bet interval of \cref{thm:bet_reg} evolves with time. Asymptotically this interval is expected to be $[0,1)$ as both the MMD and variance estimators converge to positive values while the first term in the denominator decreases to zero. That said, this interval is not necessarily monotonic, though we expect it to roughly be the case on average. We plot this interval along with the actual bet values used by FLOE (without regularization) in \cref{fig:bet_interval}. This figure makes it clear why the regularizer is necessary; initial bets are often far more aggressive than they should be.

\textbf{CIFAR-10 vs 10.1 Features.}
The featurizer we use is a ResNet-50 model \citep{he2015deepresiduallearningimage} pretrained on ImageNet and provided by the torchvision library \citep{paszke2019pytorchimperativestylehighperformance}. In particular we use the weights corresponding to ``IMAGENET1K\_V2'', and make a few modifications to the architecture: the maxpool layer is removed, and the last fully-connected layer is replaced with a linear layer that projects the number of features down to 100.

\textbf{Regularizing ONS.}
\label[appendix]{app:ons_reg}
Our regularizer not only applies to FLOE, but to any general betting strategy as long as the prediction strategy is ERM. Thus, we could also regularize ONS this way. It is worth mentioning that ONS is a ``parameter-free'' algorithm, meaning it will converge no matter the choice of learning rate. However, in finite samples these hyperparameters still play a factor. Our experiments use essentially the same hyperparameters as those of \citep{shekar2023seqmmd}, with the only difference being the choice of the inverse step size $\gamma$. Typically $\gamma$ is set to be $\gamma = 2(b - \log(1 + b)) / b^2$ where $b$ is an upper bound on $\inner{\nabla (-\log(1 + \lambda V))}{ \lambda - \lambda'}$. This ensures suitable regret bounds. For the test of \citep{shekar2023seqmmd} they clip the bets to $[-1/2,1/2]$, and for this interval $b=2$ which leads to their choice of $\gamma$. In our case the bets are in $[0,1)$ which, in the limit as $V_t\to V_\infty > 0$, gives a bound $b=1$ and consequently $\gamma = 2(1 - \log(2))$. As for the initial Hessian approximation $a_0$, we do the same as them and set it to $1.0$. 

\begin{figure}[hbt!]
  \centering
  \begin{subfigure}{\linewidth}
    \centering
    \includegraphics[width=0.4\linewidth]{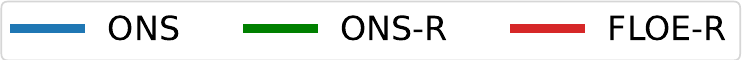}
  \end{subfigure}
  \begin{subfigure}[b]{\linewidth}
    \centering
    \includegraphics[width=\linewidth]{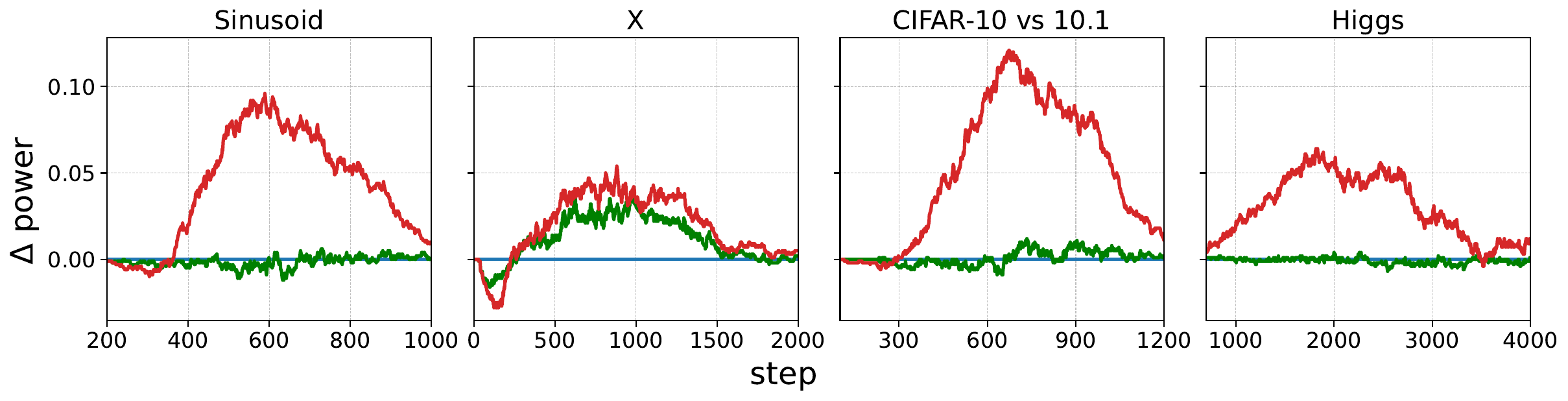}
  \end{subfigure}
  \caption{Test power of regularization strategies relative to standard ONS.}
  \label{fig:baselines_ons_rel_power}
\end{figure}

\Cref{fig:baselines_ons_rel_power} plots power of ONS with our regularizer (ONS-R) relative to the performance of ONS without the regularizer. The relative power of FLOE-R is also included for reference. The regularization strength is set with $\delta_0=\delta_1=0.1$ and $\alpha=0.2$, which is slightly weaker than what we used for our other experiments. This is because ONS seems to perform better with a weaker regularizer than FLOE, as its step sizes are naturally smaller. These smaller step sizes could also explain why regularization appears less impactful. For Sinusoid, Higgs, and CIFAR, there is essentially no difference between ONS and ONS-R; ONS-R does seem to help in the X-dataset.

\begin{figure}[hbt!]
  \centering
  \begin{subfigure}{\linewidth}
    \centering
    \includegraphics[width=0.7\linewidth]{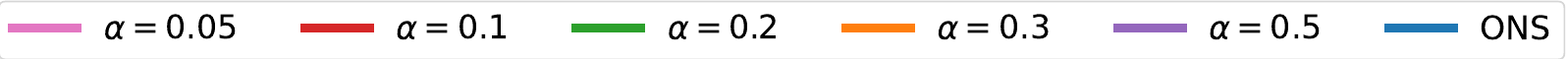}
  \end{subfigure}
  \begin{subfigure}[b]{\linewidth}
    \centering
    \includegraphics[width=\linewidth]{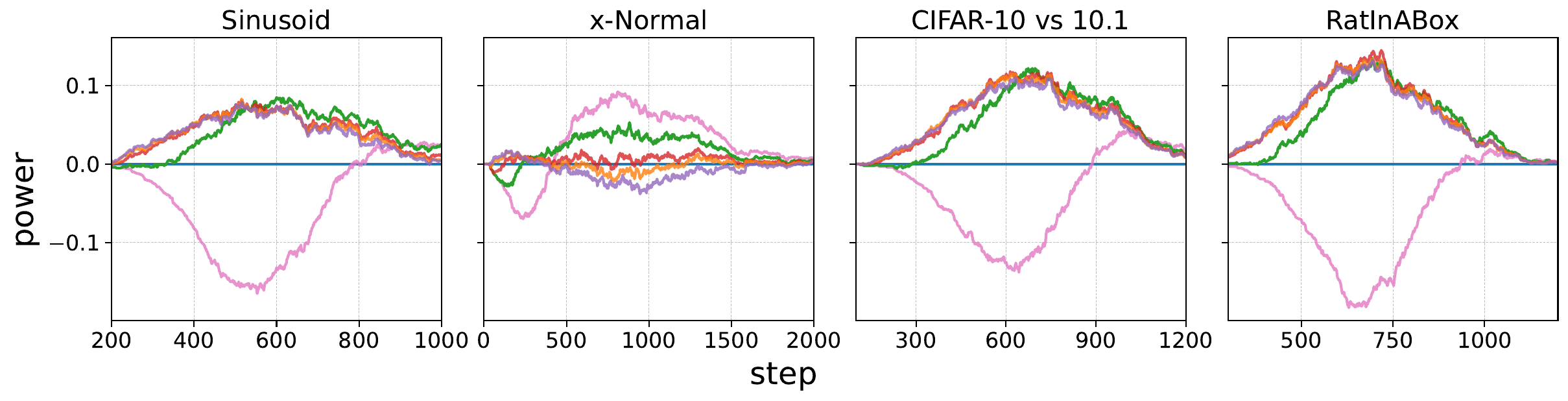}
  \end{subfigure}
  \caption{Power of FLOE-R relative to ONS at varying regularization strengths $\alpha$ and fixed $\delta = 0.1$.}
  \label{fig:sensitivity_alpha_rel_power}
\end{figure}

\textbf{Hyperparameter Sensitivity.}
\label[appendix]{app:hyper_sens}
We assess the sensitivity of the regularization parameter $\alpha$ for the practical $(\alpha, \delta_0 + \delta_1)$-regularizer constructed based on \cref{thm:bet_reg}. We set $\delta_0 = \delta_1 = 0.1$ and vary $\alpha$ between $0.05$ and $0.5$. The results are presented in \cref{fig:sensitivity_alpha_rel_power} as the power relative to ONS. In general, it seems $\alpha$ is slightly more sensitive than $\delta_0,\delta_1$ for small extremes, but still robust overall when considering values around $0.1$ and larger. Our recommendation is to keep $\alpha=\delta_0=\delta_1=0.1$ and if necessary, tweak the error rates $\delta_0$ and $\delta_1$. 

\begin{figure}[hbt!]
  \centering
    \begin{subfigure}{\linewidth}
      \centering
      \includegraphics[width=0.7\linewidth]{figures/sensitivity/delta/legend.pdf}
    \end{subfigure}
  \begin{subfigure}[b]{\linewidth}
    \centering
    \includegraphics[width=\linewidth]{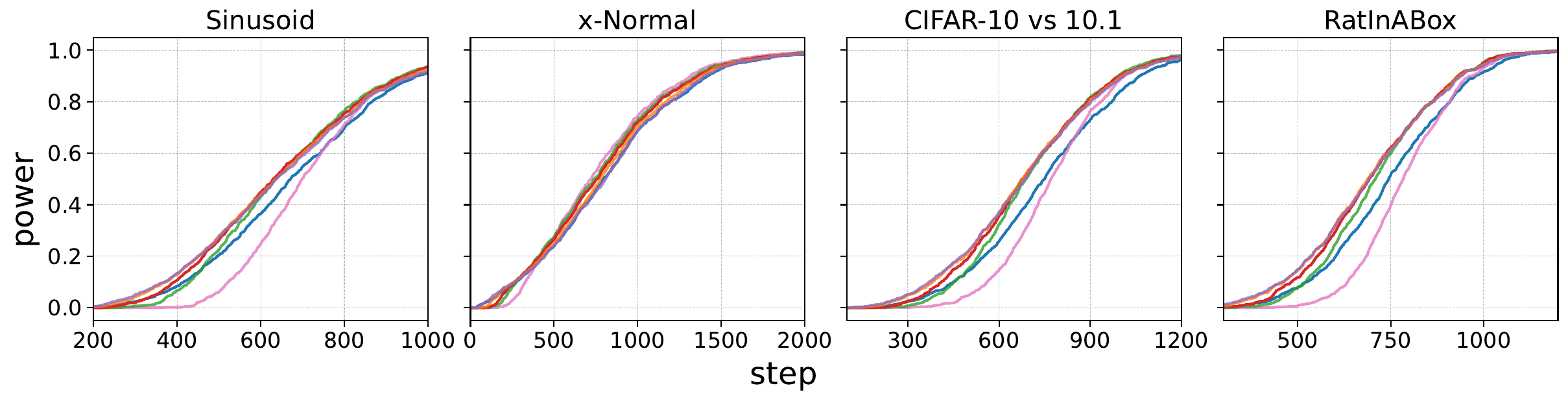}
  \end{subfigure}
  \caption{Power of FLOE-R relative to ONS at varying regularization strengths $\delta$ and fixed $\alpha=0.1$.}
  \label{fig:sensitivity_delta_raw_power}
\end{figure}

For completeness, we also include in \cref{fig:sensitivity_delta_raw_power} the raw power plots corresponding to the relative power plots of \cref{fig:sensitivity_delta_rel_power}.

\textbf{Kernel Bandwidth.}
All our experiments use the Gaussian kernel $k(x,y) = \exp(-\frac{1}{2}\|x-y\|^2 / \gamma^2)$ at problem-specific bandwidths; these bandwidths were selected to approximately maximize the power of the \textit{ONS betting strategy} (rather than FLOE or FLOE-R) on a validation set. This mean power was computed over 100 tests at time $t=1000$ for the x-Normal, CIFAR, and RatInABox problems, time $t=800$ for the Sinusoid and Blobs problems, and $t=3500$ for the Higgs dataset. This results in the following bandwidths:
the Sinusoid problem uses $\gamma=0.3$; the Blobs dataset uses $\gamma = 0.3$; the x-Normal and CIFAR datasets uses $\gamma = 0.5$; the RatInABox dataset uses $\gamma=0.1$; the Higgs dataset uses $\gamma = 0.4$. The same procedure was used for the sequential kernel independence test, yielding $\gamma_X = \gamma_Y = 0.18$ for Sinusoid, $\gamma_X = \gamma_Y = 0.9$ for x-Normal, and $\gamma_X =0.35, \gamma_Y = 0.63$ for RatInABox.

\begin{figure}[hbt!]
  \centering
    \begin{subfigure}{\linewidth}
      \centering
      \includegraphics[width=0.4\linewidth]{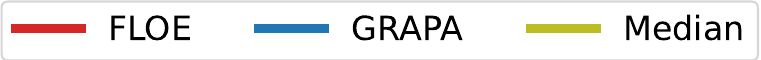}
    \end{subfigure}
  \begin{subfigure}[b]{\linewidth}
    \centering
    \includegraphics[width=\linewidth]{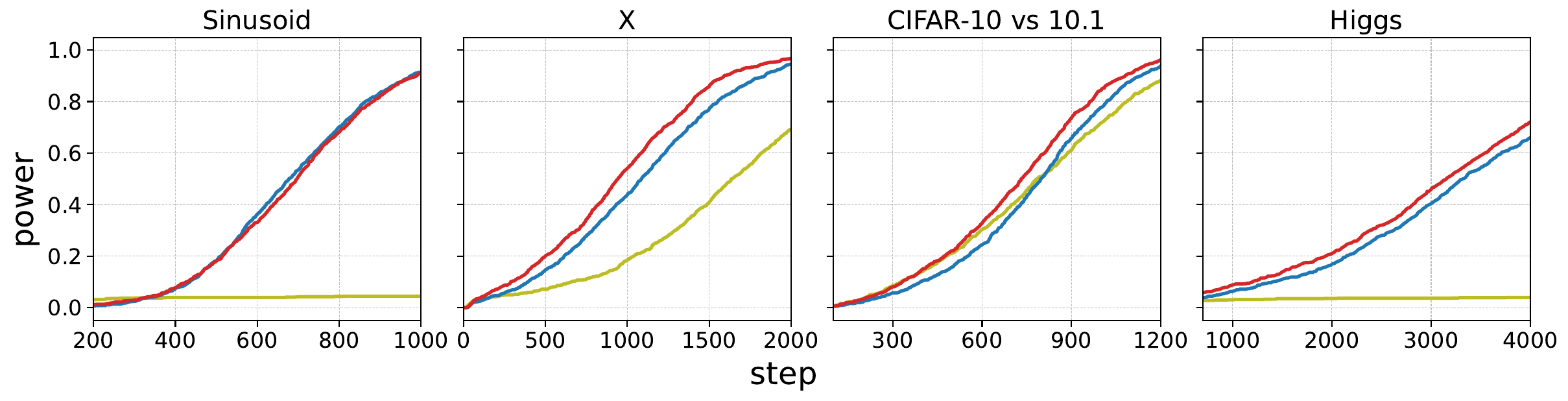}
  \end{subfigure}
  \caption{Kernel selection strategies on a variety of problems.}
  \label{fig:kernel_selection}
\end{figure}

\textbf{Kernel Selection.}
Since the only requirement on the prediction strategy is that it must be predictable, we can select the kernel bandwidth of the critic at time $t$ based on any criterion computed from the samples up to time $t-1$. One approach is to select the bandwidth that maximizes the wealth at time $t-1$; this is the same as GRAPA except applied to the model parameters instead of the bets. Alternatively, we could also optimize the e-power of the current step, which can be done by using the loo e-power estimates, i.e. 
\begin{align*}
    \gamma_t = \argmax_\gamma \sum_{\ell=1}^{t-1} \log( 1 + \lambda_t^{(\gamma)} V_\ell^{(t-1, \gamma)} ).
\end{align*}
This strategy is essentially FLOE applied to the kernel bandwidths $\gamma$.
\Cref{fig:kernel_selection} compares the GRAPA and FLOE kernel selection approaches with the median heuristic; the reported power is averaged over 1,000 tests and each baseline uses the FLOE betting strategy. To speed up the optimization, we consider a discrete set of 20 evenly-spaced bandwidths between $[0.1,0.7]$ so that each e-power estimate can be computed in parallel.
Overall, FLOE outperforms GRAPA in most problems. Even in Sinusoid, its worst performing problem, FLOE is only slightly worse.

\begin{figure}[hbt!]
  \centering
  \begin{subfigure}[b]{\linewidth}
    \centering
    \includegraphics[width=\linewidth]{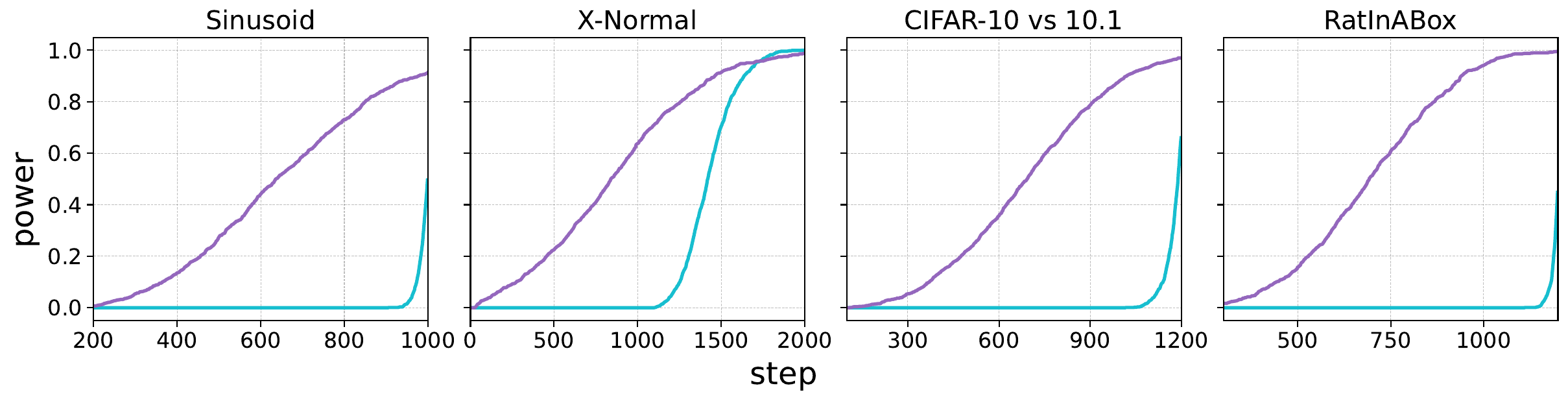}
  \end{subfigure}
  \caption{Power of FLOE (purple) against STaR-Bets (neon blue).}
  \label{fig:star_bets}
\end{figure}

\textbf{Performance against STaR-Bets}.
As an additional baseline, we compare FLOE to the STaR-Bets betting strategy of \citet{voráček2025starbetssequentialtargetrecalculatingbets} for the bounded mean testing problem. To translate their approach (Algorithm 4) to our kernel two-sample testing process, we replace their payoffs $X_t - m$ with our kernel critic payoffs $f_t(X_t) - f_t(X_t)$ and use $\alpha=\delta=0.05$. Results are reported in \cref{fig:star_bets}.

We find that FLOE significantly outperforms STaR-Bets for the two-sample problem. STaR-Bets' burn-in period is quite long, and by the time it starts accumulating wealth FLOE already has near perfect power. That said, after burn-in STaR-Bets' power increases at a far more rapid rate than FLOE. We also point out that FLOE does not apply to the bounded mean testing problem, as there is no critic function to be learned. It's payoffs are simply the centred samples $X_t - m$. If we were to define $f_t(X_t) = X_t - m$, FLOE would simply reduce to GRAPA.

\textbf{Assets and Licenses.}
We describe some of the assets and their licenses used in our work:
\begin{itemize}
    \item Higgs Dataset \citep{baldi2014higgs} available from the UCI Machine Learning Repository under the CC BY 4.0 license.
    \item CIFAR-10 \citep{krizhevsky2009learning} distributed via the UCI Machine Learning Repository under CC BY 4.0.
    \item CIFAR-10.1 \citep{recht2018cifar10classifiersgeneralizecifar10} released under the MIT license.
    \item A pretrained ResNet-50 model \citep{he2015deepresiduallearningimage} provided by Torchvision \citep{paszke2019pytorchimperativestylehighperformance} and released under the BSD-3-Clause license.
\end{itemize}

\makeatletter
\if@preprint\else
\newpage
\input{checklist.tex}
\fi
\makeatother

\end{document}

%% file: checklist.tex
\section*{NeurIPS Paper Checklist}

\begin{enumerate}

\item {\bf Claims}
    \item[] Question: Do the main claims made in the abstract and introduction accurately reflect the paper's contributions and scope?
    \item[] Answer: \answerYes{} %
    \item[] Justification: We touch on our main contributions in the abstract, and provide more details in the introduction (particularly the sub-section titled "contributions").

\item {\bf Limitations}
    \item[] Question: Does the paper discuss the limitations of the work performed by the authors?
    \item[] Answer: \answerYes{} %
    \item[] Justification: The "practical considerations" of section 3.3 discusses limitations of our regularization strategy and proposes a possible workaround. Section 4 also touches on a potential situation where including this regularization is not ideal, as well as including ablation on the sensitivity of the regularizer hyperparameters. Computation complexity is briefly discussed in Section 3.2 and 3.3.

\item {\bf Theory assumptions and proofs}
    \item[] Question: For each theoretical result, does the paper provide the full set of assumptions and a complete (and correct) proof?
    \item[] Answer: \answerYes{} %
    \item[] Justification: Full proofs for each theorem are provided in the appendix and, to the best of our knowledge, are correct.

    \item {\bf Experimental result reproducibility}
    \item[] Question: Does the paper fully disclose all the information needed to reproduce the main experimental results of the paper to the extent that it affects the main claims and/or conclusions of the paper (regardless of whether the code and data are provided or not)?
    \item[] Answer: \answerYes{} %
    \item[] Justification: Our proposed test procedure is described in detail and, apart from the inherent randomness of the data, all dataset settings are described in Section 4 and Appendix B.

\item {\bf Open access to data and code}
    \item[] Question: Does the paper provide open access to the data and code, with sufficient instructions to faithfully reproduce the main experimental results, as described in supplemental material?
    \item[] Answer: \answerYes{} %
    \item[] Justification: The code we used for all experiments is provided as supplementary material. However, only partial instructions are currently given.

\item {\bf Experimental setting/details}
    \item[] Question: Does the paper specify all the training and test details (e.g., data splits, hyperparameters, how they were chosen, type of optimizer) necessary to understand the results?
    \item[] Answer: \answerYes{} %
    \item[] Justification: All hyperparameters pertaining to our datasets and method are described in section 4 and Appendix B.

\item {\bf Experiment statistical significance}
    \item[] Question: Does the paper report error bars suitably and correctly defined or other appropriate information about the statistical significance of the experiments?
    \item[] Answer: \answerNo{} %
    \item[] Justification: We considered plotting empirical confidence intervals on the power plots of our experiments but found this made results harder to read and were not very informative. Each of our experiments are run 1,000 times and so the mean power is relatively accurate.

\item {\bf Experiments compute resources}
    \item[] Question: For each experiment, does the paper provide sufficient information on the computer resources (type of compute workers, memory, time of execution) needed to reproduce the experiments?
    \item[] Answer: \answerNo{} %
    \item[] Justification: The compute resources used is not imperative to our approach. A GPU is helpful, but not required, and memory usage is small enough so that an everyday computer can run it.
    
\item {\bf Code of ethics}
    \item[] Question: Does the research conducted in the paper conform, in every respect, with the NeurIPS Code of Ethics \url{https://neurips.cc/public/EthicsGuidelines}?
    \item[] Answer: \answerYes{} %
    \item[] Justification: Our research conforms to the NeurIPS code of ethics.

\item {\bf Broader impacts}
    \item[] Question: Does the paper discuss both potential positive societal impacts and negative societal impacts of the work performed?
    \item[] Answer: \answerNo{} %
    \item[] Justification: In the introduction, we briefly discussed positive societal impacts in terms of mitigating p-hacking. We did not address any negative societal impacts.
    
\item {\bf Safeguards}
    \item[] Question: Does the paper describe safeguards that have been put in place for responsible release of data or models that have a high risk for misuse (e.g., pre-trained language models, image generators, or scraped datasets)?
    \item[] Answer: \answerNA{} %
    \item[] Justification: There are no such risks.

\item {\bf Licenses for existing assets}
    \item[] Question: Are the creators or original owners of assets (e.g., code, data, models), used in the paper, properly credited and are the license and terms of use explicitly mentioned and properly respected?
    \item[] Answer: \answerYes{} %
    \item[] Justification: We have included a section in Appendix B detailing the assets and licenses used.

\item {\bf New assets}
    \item[] Question: Are new assets introduced in the paper well documented and is the documentation provided alongside the assets?
    \item[] Answer: \answerNA{} %
    \item[] Justification: No new assets are introduced.

\item {\bf Crowdsourcing and research with human subjects}
    \item[] Question: For crowdsourcing experiments and research with human subjects, does the paper include the full text of instructions given to participants and screenshots, if applicable, as well as details about compensation (if any)? 
    \item[] Answer: \answerNA{} %
    \item[] Justification: No crowdsourcing or human-based experiments are used.

\item {\bf Institutional review board (IRB) approvals or equivalent for research with human subjects}
    \item[] Question: Does the paper describe potential risks incurred by study participants, whether such risks were disclosed to the subjects, and whether Institutional Review Board (IRB) approvals (or an equivalent approval/review based on the requirements of your country or institution) were obtained?
    \item[] Answer: \answerNA{} %
    \item[] Justification: No crowdsourcing or human-based experiments are used.

\item {\bf Declaration of LLM usage}
    \item[] Question: Does the paper describe the usage of LLMs if it is an important, original, or non-standard component of the core methods in this research? Note that if the LLM is used only for writing, editing, or formatting purposes and does \emph{not} impact the core methodology, scientific rigor, or originality of the research, declaration is not required.
    \item[] Answer: \answerNA{} %
    \item[] Justification: Our proposed method was not developed with LLM use.

\end{enumerate}